%% file: acl_latex.tex
\documentclass[11pt]{article}

\usepackage[preprint]{acl}

\usepackage{times}
\usepackage{latexsym}

\usepackage[T1]{fontenc}

\usepackage[utf8]{inputenc}

\usepackage{microtype}

\usepackage{inconsolata}

\usepackage{graphicx}

\usepackage{multirow}
\usepackage{booktabs}
\usepackage{amsmath}
\usepackage{amsfonts}
\usepackage{amssymb}
\usepackage{subcaption}
\usepackage{threeparttable}
\usepackage{algorithm}
\usepackage{algorithmic}
\usepackage{makecell}

\usepackage{colortbl}
\usepackage{xcolor}
\newcommand{\red}[1]{\textcolor{red}{#1}}

\usepackage{xcolor,colortbl}
\usepackage{pgfmath}
\usepackage{pgf}

\usepackage[table]{xcolor}
\definecolor{FreshBlue}{RGB}{173,216,230}

\definecolor{heatblue}{RGB}{218,227,243}   
\definecolor{heatred}{RGB}{244,209,209}     
\newcommand{\posheat}[2]{\cellcolor{white!#1!heatblue}{#2}}
\newcommand{\negheat}[2]{\cellcolor{white!#1!heatred}{#2}}
\newcommand{\eqheat}[1]{#1}

\newcommand{\best}[1]{\textbf{#1}}
\newcommand{\second}[1]{\underline{#1}}

\usepackage{textcomp}
\newcommand{\showslash}{\texttt{\red\textbackslash}}
\newcommand{\showspace}{\red{\textvisiblespace}}
\newcommand{\shownl}{\red{\texttt{\textbackslash n}}}
\newcommand{\showlb}{\red{\texttt{\textbackslash[}}}
\newcommand{\showrb}{\red{\texttt{\textbackslash]}}}
\newcommand{\cellsmall}{\setlength{\parskip}{2pt}\setlength{\parindent}{0pt}}

\newcommand{\redtexttt}[1]{\texttt{\textcolor{red}{#1}}}

\usepackage[most]{tcolorbox}
\definecolor{colframecolor}{RGB}{55,98,175}   
\definecolor{colbackcolor}{RGB}{218,227,243}  

\usepackage{pifont}

\usepackage[misc]{ifsym}

\usepackage{fontawesome5}

\definecolor{nred}{RGB}{196, 38, 11}
\definecolor{ngreen}{RGB}{18, 141, 21}
\definecolor{nblue}{RGB}{41, 52, 190}
\definecolor{hzw}{RGB}{223, 97, 76}
\definecolor{lt}{RGB}{54, 89, 170}
\definecolor{tblue}{rgb}{0.867, 0.922, 0.969}
\definecolor{zlblue}{RGB}{196, 223, 251}

\title{RLVR Datasets and Where to Find Them: \\Tracing Data Lineage for Better Training Data}

\author{
    Hsiu-Yuan Huang$^{1,2,3}$,
    Weijie Liu$^{3}$\thanks{~~Corresponding author.},
    Chenming Tang$^{1,2}$,
    Sanwoo Lee$^{1,2}$,
    \\
    \textbf{Kai Yang}$^{3}$,
    \textbf{Yangkun Chen}$^{3}$,
    \textbf{Saiyong Yang$^{3}$,}
    \textbf{Yunfang Wu$^{1,2}$\footnotemark[1]}
    \\
    $^{1}$National Key Laboratory for Multimedia Information Processing, Peking University \\ 
    $^{2}$School of Computer Science, Peking University \\
    $^{3}$LLM Department, Tencent \\
    \small{ \Letter \; \href{mailto:huang.hsiuyuan}{huang.hsiuyuan@stu.pku.edu.cn}, \href{mailto:wuyf@pku.edu.cn}{wuyf@pku.edu.cn} }\\
}

\begin{document}
\maketitle

\input{section/0_abstract}
\input{section/1_introduction}

\input{section/2_data_lineage}

\input{section/3_benchmark_of_dataset}

\input{section/4_data_curation}
\input{section/5_experiments}

\input{section/6_conclusion}
\input{section/limitations}

\section*{Ethics Statement}
\paragraph{Use of AI Assistants.}
We have employed ChatGPT as a writing assistant, primarily for polishing the text after the initial composition.
We certify that any use of AI tools, including ChatGPT, was strictly limited to linguistic refinement such as improving grammar, clarity, and style. All substantive ideas, analyses, and arguments presented in this work originate from the authors or from properly cited prior research.

\paragraph{Computational Budget.}
All our experiments are conducted on a machine with CentOS 8, 384 AMD$^\circledR$ EPYC\texttrademark{} 9K84 96-Core Processor CPUs and 2.2TiB memory. 
We use 8$\times$ NVIDIA H20 GPUs for all experiments. GRPO training takes approximately 3 and 7 days per experiment for the 1.7B and 8B models, respectively.

\paragraph{Reproducibility.}
Our work is reproducible because we have provided our source code and implementation details.

\paragraph{Scientific Artifacts.}
The scientific artifacts we use are
shown in Table~\ref{tab:dataset_details}, with links provided.
License information can be found at the corresponding links. All existing artifacts were used in a manner consistent with their intended purpose, namely RLVR training and evaluation. All datasets used and constructed in this work are derived from publicly available open-source data; therefore, data consent considerations are not applicable to this work.

\paragraph{Clarification on Human Subjects.}
The data annotation and verification procedures in this work were conducted entirely by the authors through case-by-case manual inspection. \textbf{No} crowdworkers, external annotators, or paid labeling services were recruited during the annotation process. Therefore, considerations regarding crowdworker recruitment and compensation are not applicable to this work.

\paragraph{Potential Risks.}
To the best of our knowledge, there are no potential risks concerning our work.



\bibliography{custom}

\clearpage
\appendix
\input{section/appendix}

\end{document}

%% file: section/0_abstract.tex
\begin{abstract}

The proliferation of Reinforcement Learning from Verifiable Rewards (RLVR) datasets has exacerbated provenance collapse due to unclear lineage among existing datasets. 
To bridge this fragmented RLVR data landscape, we propose Atomic-source Tracing via Lineage-Aware Search (ATLAS), a systematic framework for tracing RLVR datasets back to their atomic sources, attributing over 99.7\% of 1.45M instances to 20 atomic sources. Our analysis reveals that most RLVR datasets are variants of a small set of shared upstream sources, with few introducing genuinely new data, and many facing data contamination risks. These findings naturally motivate us to curate a new RLVR dataset, DAPO++, and to benchmark existing datasets from a lineage-aware perspective.
To this end, we propose Source-level Counterfactual Attribution (SCA) as a guiding principle to curate a decontaminated training dataset with concentrated learning signals. Essentially, SCA measures a sample's marginal utility by comparing per-atomic-source RL checkpoints against a shared base model. 
Building upon these attribution signals, we further design a composite dataset quality score $Q$ that strongly correlates with downstream RLVR performance. Experiments on Qwen3 series models verify that DAPO++ consistently improves performance on held-out benchmarks, while $Q$ reliably predicts downstream RLVR training effectiveness. Our code and data is available at 
\url{https://github.com/Celine-hxy/ATLAS}.

\end{abstract}

%% file: section/1_introduction.tex
\section{Introduction}
\emph{\quad Where Do We Come From? What Are We? Where Are We Going?}   \quad \quad \quad \; \; \; \; \; \; \; \;---Paul Gauguin

\begin{figure}[htbp]
  \centering
  \includegraphics[width=0.9\linewidth]{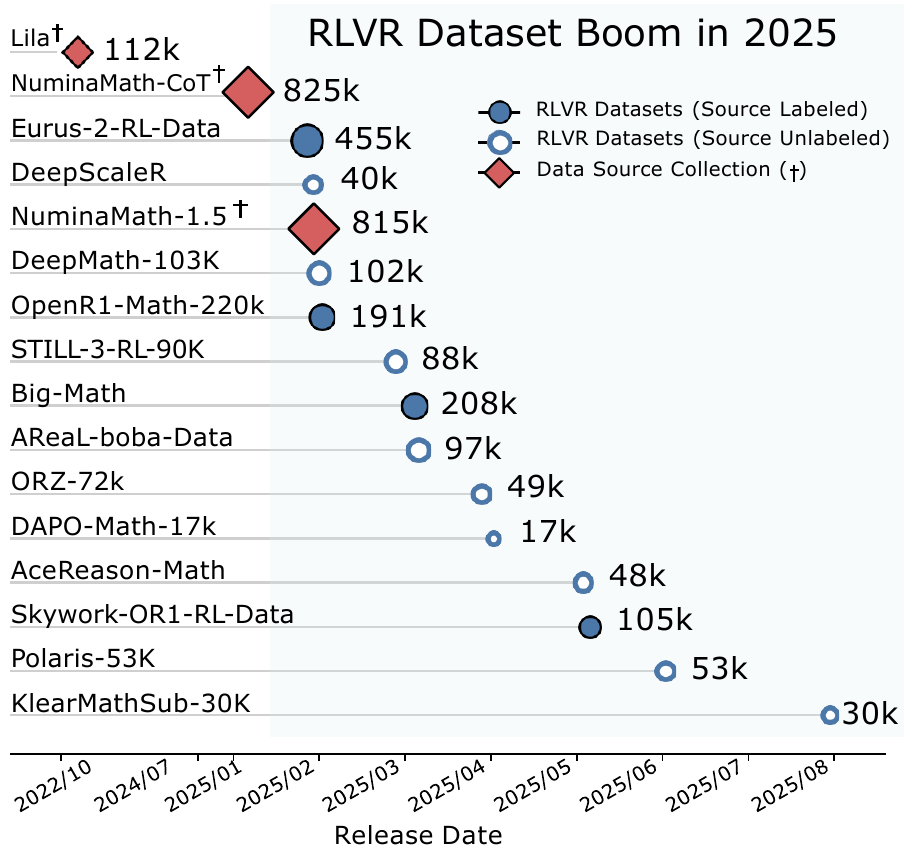}
  \caption{Timeline of RLVR dataset releases. The field has grown rapidly since 2025, with many datasets being ``\textbf{\textit{openly-closed}}'', meaning that although the datasets are publicly available, their underlying data provenance remains opaque.}
  \label{fig:teaser_data_source}
\end{figure}

The emergence of Reinforcement Learning from Verifiable Rewards (RLVR) has driven rapid growth in both RLVR research and training corpora. Existing studies have investigated RLVR from various perspectives, including model behavior, optimization dynamics, and data attribution~\cite{luffy2025,zhou2026efficientrlvrtrainingweighted}. However, these works are often trained and evaluated on substantially different datasets, making empirical results difficult to compare fairly. Meanwhile, the rapid proliferation of RLVR datasets (Figure~\ref{fig:teaser_data_source}), many of which lack clear provenance or source annotation, has raised a practical yet underexplored question: \textbf{\textit{where do these RLVR datasets come from, and how can we find better datasets for RLVR training?}}

One of our goals is to trace data back to its \textit{atomic sources}, namely the original points of collection or construction (e.g., a specific book, website, competition, or crowdsourced effort), rather than merely tracing datasets back to their immediate predecessors. This is motivated by our finding that modern RLVR datasets are often built upon prior datasets through filtering, rewriting, recomposition, and iterative aggregation, resulting in substantial overlap across datasets (\S~\ref{sec:data_landscape}). We argue that atomic-source tracing provides the following perks:
\textbf{(1) ``Where Do We Come From''---tracing errors, synthetic data, and leakage:} data lineage tracing improves dataset transparency by disentangling the hidden composition of \textit{openly-closed} datasets (Figure~\ref{fig:teaser_data_source}), making it possible to address \textit{provenance collapse}~\cite{brooks2025synthetic}, namely the loss of information about where data came from, who modified it, and whether it can be trusted.
\textbf{(2) ``What Are We''---encouraging quality-oriented data development:} provenance tracing reduces repetitive data cleaning and saves human and computational resources, enabling greater focus on high-quality data construction.
\textbf{(3) ``Where Are We Going''---exploring uncharted data sources:} understanding existing data provenance helps reveal which domains remain underexplored, providing guidance for future data mining.
To address these, we propose an \textit{\textbf{A}tomic-source \textbf{T}racing via \textbf{L}ineage-\textbf{A}ware \textbf{S}earch} (ATLAS) framework (\S~\ref{sec:lineage_tracing_framework}), through which we trace 1,450,827 RLVR instances back to 20 atomic data sources and provide detailed analysis (\S~\ref{sec:data_landscape}), including million-scale pairwise similarity matching against 14 mathematical evaluation benchmarks to detect leakage, uncovering 36,148 leaked instances across datasets. Insights derived from ATLAS further motivate us to curate a new RLVR dataset \textsc{DAPO++}.

Another goal of this work is to systematically benchmark existing RLVR datasets, thereby providing a roadmap for the dataset construction and selection, both of which constitute a fundamental prerequisite for RLVR research. 
Built upon ATLAS, we further propose \textit{\textbf{S}ource-level \textbf{C}ounterfactual \textbf{A}ttribution} (SCA), a source-aware attribution framework that enables learnability annotation and source-level utility estimation (\S~\ref{sec:SCA}). Together, ATLAS and SCA form the foundation of our RLVR dataset benchmarking and scoring framework (\S~\ref{sec:rlvr_score}), enabling systematic ranking and analysis of existing RLVR datasets.

Our contributions can be summarized as follows:
\begin{itemize}
    \item We propose ATLAS, a data lineage tracing framework for RLVR datasets, and provide detailed analyses of dataset provenance and benchmark leakage, and curate a new dataset \textsc{DAPO++}, based on the insights derived from our analysis.

    \item Built upon ATLAS, we propose \textbf{Source-level Counterfactual Attribution (SCA)}, a source-aware attribution method that estimates the contribution of atomic data sources to RLVR training effectiveness while providing fine-grained instance-level learnability annotations.

    \item Leveraging ATLAS and SCA, we systematically benchmark existing RLVR datasets, to support future RLVR training and dataset construction.
\end{itemize}

%% file: section/2_data_lineage.tex
\section{RLVR Dataset Lineage and Provenance}
\label{sec:dataset_lineage_provenance}
To facilitate the discussion of our \textbf{A}tomic-source \textbf{T}racing via \textbf{L}ineage-\textbf{A}ware \textbf{S}earch (\textbf{ATLAS}) framework for RLVR training corpora, we first clarify the key concepts and terminology involved.
\begin{itemize}
    \item \textbf{Data Source:} A collection of instances not intended for RLVR training. If its origin is singular or its data follow a consistent standard, it is referred to as an \textit{atomic source}.
    \item \textbf{Dataset:} A general term for a collection of data that may aggregate instances, data sources, or other datasets. In this paper, it mostly refers to the data collection associated with a HuggingFace dataset identifier.
    \item \textbf{RLVR Dataset:} A dataset specifically constructed for RLVR training, whose contents are typically composition of multiple data sources.
\end{itemize}

\begin{figure*}[htbp]
  \centering
  \includegraphics[width=\linewidth]{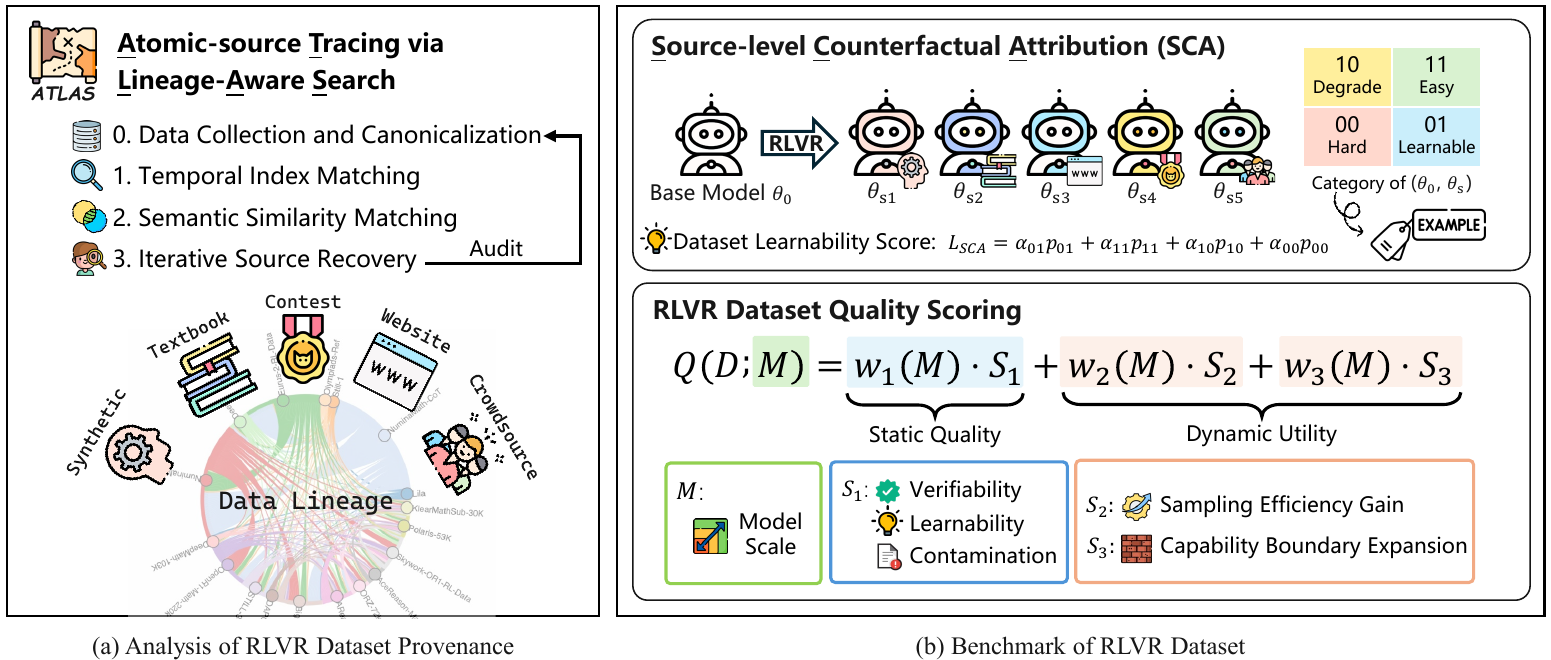}
  \caption{Overview of our framework for RLVR dataset provenance analysis and benchmarking.
  }
  \label{fig:main}
\end{figure*}

\subsection{ATLAS Framework}
\label{sec:lineage_tracing_framework}
Our ATLAS framework is illustrated in Figure~\ref{fig:main}(a), while the corresponding pseudocode is presented in Appendix~\ref{app:data_lineage_tracing}. And detailed discussions of related work are provided in the Appendix \ref{app:related_work}, due to space limitations.

\paragraph{Stage 0: Data Collection and Canonicalization.}
We collect a set of commonly used RLVR datasets from the open-source community, along with the upstream datasets on which they are built, forming the initial seed dataset pool for source tracing.
Detailed descriptions of the manually collected and thoroughly inspected datasets, which constitute the core infrastructure of our study, are provided in Table~\ref{tab:dataset_details}, including release dates, Hugging Face IDs, dataset sizes, associated papers, and Hugging Face links.

Due to substantial heterogeneity across datasets, including schema, storage formats, and prompt structures, data canonicalization is non-trivial and typically requires manual inspection. For example, a single dataset may append multiple instructions of varying formats after each prompt. To address this, we manually inspect each dataset by sampling 30 to 50 instances, from which we extract and standardize the final question-answer pairs.

\paragraph{Stage 1: Temporal Index Matching.}
To efficiently determine the occurrence status of a given prompt, we design a hash-based temporal indexing mechanism that avoids costly string-level matching. For each prompt $p$, we compute a 40-character hexadecimal identifier $h$ using \textit{SHA-1} hashing, enabling efficient lookup across datasets.
We then traverse all instances in chronological order while maintaining a global lineage dictionary $\mathcal{L}[h]$ which stores instance metadata together with the occurrence list $\mathcal{O}[h]$, and update the record accordingly with metadata when provided.

\paragraph{Stage 2: Semantic Similarity Matching.}
To recover source attribution for instances unresolved by exact matching, we perform semantic similarity matching over the unmatched set $\mathcal{U}$, where mismatches often arise from prompt-level transformations.
Specifically, we encode the entire corpus using Sentence-BERT embeddings~\cite{sentencebert-2019} and trace unmatched instances back to their potential source data through cosine similarity retrieval under human auditing, in which we manually verify them case-by-case and judge whether they match.

\paragraph{Stage 3: Iterative Source Recovery.}
We manually inspect their question types and formatting patterns to identify potential missing sources, if the number of unmatched instances in $\mathcal{U}$ exceeds a threshold $\tau$. Based on these observations, we select candidate datasets $\mathcal{D}_{\text{new}}$ from potential repositories, primarily authoritative releases not yet included in data sources, incorporate them into the global dataset pool, and rerun Stage~0--2 matching.
Subsequently, candidate datasets $\mathcal{D}_{\text{new}}$ that fail to recover meaningful matches are discarded from the dataset pool.
This iterative process continues until the number of unmatched instances in $\mathcal{U}$ falls below $\tau$. The remaining unmatched instances are labeled as \textit{unknown}. 

\subsection{RLVR Dataset Provenance}
\label{sec:data_landscape}
In our implementation, we set the $\tau=1\%$, and the final lineage dictionary contains 1,450,827 instances, with fewer than 1\% labeled as \textit{unknown}.

\begin{figure*}[t]
\centering
  \includegraphics[width=\linewidth]{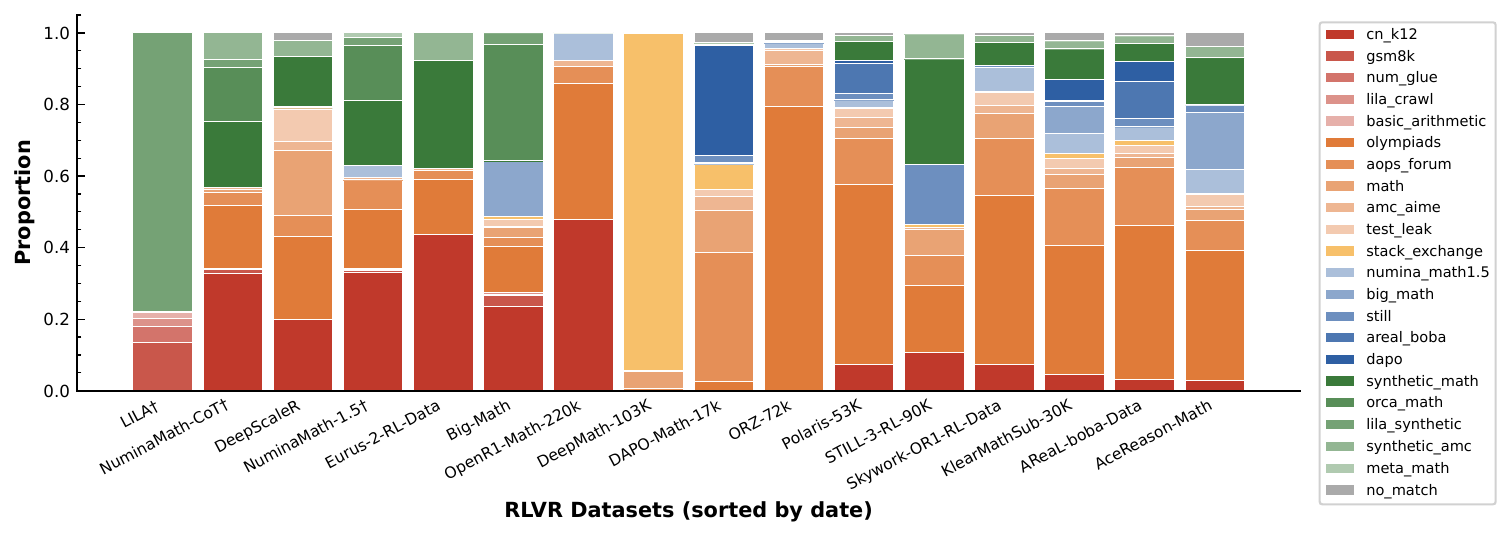}
  \caption{
  Atomic-source composition and relative proportions. Major source datasets are marked with $\dagger$. We manually annotate each atomic source with a problem-type label and group them into several color-coded categories, as detailed in Table~\ref{tab:data_source_detail}. Within each type, larger atomic sources are represented using darker shades.
}
  \label{fig:source_stackbar}
\end{figure*}

\paragraph{Source Decomposition.}
Figure~\ref{fig:source_stackbar} presents the atomic-source decomposition of representative RLVR training datasets traced by our ATLAS framework, while Figure~\ref{fig:source_heatmap} (Appendix~\ref{app:rlvr_provenance}) further reveals the specific RLVR datasets in which these atomic data sources were first introduced. Although many datasets exhibit substantial atomic source overlap,
we still observe considerable heterogeneity in the relative source proportions across datasets. 

\paragraph{Dataset Contribution.}
We further visualize the compositional relationships among RLVR datasets using Bokeh plots, as shown in Figure~\ref{fig:main}(a), where the detailed views for each dataset are provided in Appendix~\ref{app:rlvr_provenance}. 
Moreover, Eurus-2-RL-Data makes a notable contribution by extensively cleaning and transforming the MCQs in NuminaMath-CoT into open-ended reasoning formats, which were proven effective (\S~\ref{sec:result}) and were subsequently inherited by a large number of later RLVR datasets.

\paragraph{Dataset Leakage Risk Analysis}
\label{sec:dataset_leak}

\begin{figure*}[t]
\centering
  \includegraphics[width=\linewidth]{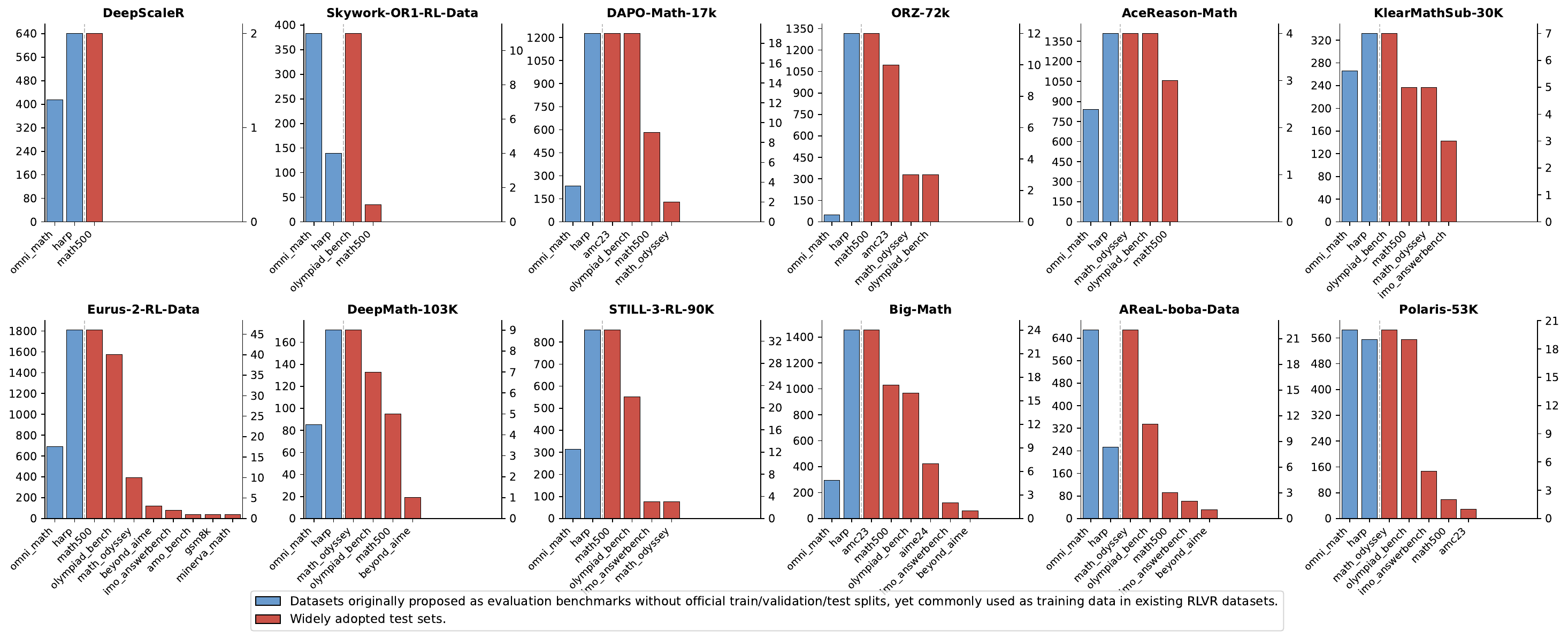}
  \caption{Leakage severity of RLVR datasets measured by semantic similarity to benchmark test sets.
  }
  \label{fig:leak_per_rlvr_dataset}
\end{figure*}

We conduct large-scale \textbf{\textit{exhaustive pairwise similarity matching}} between RLVR datasets and commonly used evaluation benchmarks to analyze potential benchmark leakage between existing RLVR datasets (Figure~\ref{fig:leak_per_rlvr_dataset}), major source datasets (Appendix~\ref{app:leakage_analysis}), and commonly used evaluation benchmarks. Our analysis reveals three major concerns, with detailed cases provided in Appendix~\ref{app:leakage_analysis_case}.
(1) We identify several RLVR datasets that directly include benchmarks originally proposed for RLVR evaluation as part of their training data, such as Omni-Math and HARP, leading to explicit contamination.
(2) Through similarity-based matching, we find that benchmark leakage is pervasive across current RLVR datasets. In many cases, the leaked instances differ from benchmark test samples only in superficial formatting variations (Case 1, 2 and 3), making such leakage difficult to detect through conventional rule-based filtering. 
(3) We further observe cases where benchmark instances are lightly modified or partially rewritten while still retaining extremely high semantic overlap with the original test problems (Case 4), posing substantial hidden leakage risks.

It should be emphasized that results in Figure~\ref{fig:leak_per_rlvr_dataset} only cover high-confidence leakage cases ($\mathrm{similarity} \geq 90\%$).
We nevertheless observe substantial leakage at lower similarity levels (e.g., $\mathrm{similarity} \approx 80\%$), particularly for MCQ-transformed variants with nearly identical underlying content, suggesting that the reported results may still underestimate the true extent of benchmark contamination. 
This calls for a more rigorous provenance tracking and leakage auditing, and further raise an important question: 
\textbf{\textit{how does contamination affect the utility and downstream performance of RLVR datasets?}}

%% file: section/3_benchmark_of_dataset.tex
\section{Benchmarking RLVR Datasets}

\label{sec:benchmark_rlvr_dataset}

In this section, we address the above question by proposing a multi-dimensional dataset quality scoring framework for benchmarking RLVR datasets, as illustrated in Figure~\ref{fig:main}(b). We first introduce \textit{\textbf{S}ource-level \textbf{C}ounterfactual \textbf{A}ttribution} (SCA), a data attribution method built upon the atomic sources identified by ATLAS. SCA enables instance-level learnability labeling, which subsequently serves as a key component of our scoring framework. 


\subsection{Source-level Counterfactual Attribution}
\label{sec:SCA}
A central challenge in evaluating RLVR training data is that the contribution of an individual instance cannot be directly isolated, as RL training is a global optimization process where the effect of each sample is heavily entangled with the overall training distribution~\cite{koh2020understandingblackboxpredictionsinfluence,ghorbani19_datashapley,hu2025snapshotinfluencelocaldata}. To address this, we propose SCA, a data attribution method that operates at the granularity of atomic sources from ATLAS (\S~\ref{sec:lineage_tracing_framework}). 
Specifically, SCA treats data instances from the same atomic source, whose shared origins and annotation protocols make them more likely to exhibit similar stylistic and domain characteristics, as a single intervention unit to reduce attribution confounding.
For each atomic source $s$, we train a corresponding RL checkpoint $\theta_s$, initialized from the base model $\theta_0$, thereby forming a counterfactual pair $(\theta_0,\theta_s)$. By evaluating instances $i \in s$ on both $\theta_0$ and $\theta_s$, the resulting behavioral difference provides a minimally confounded proxy for estimating the contribution of individual instances to RLVR training. 

We further perform a one-time binary labeling by comparing the correctness outcomes of $(\theta_0,\theta_s)$ on each instance, yielding four categories $(00, 01, 10, 11)$ that characterize different levels of instance learnability.
Specifically, $00$ represents unsolvable cases where fail on both the counterfactual pair, $01$ represents genuinely learnable instances where succeeds after RL, $10$ represents degrade cases, and $11$ represents overly-easy instances that is mastered by both models. Accordingly, we define the learnability score of a datase:
\begin{equation}
\small
    L_\mathrm{SCA} \;=\;
        \alpha_{01}\,p_{01}
        + \alpha_{11}\,p_{11}
        + \alpha_{10}\,p_{10}
        + \alpha_{00}\,p_{00}.
\end{equation}
where $p$ denotes the proportion of instances assigned to behavioral category, and $\alpha$ represents the corresponding learnability weight for that category.


\subsection{RLVR Dataset Quality Scoring}
\label{sec:rlvr_score}


Grounded in the two complementary perspectives of \textbf{Static Quality} and \textbf{Dynamic Utility}, whose detailed motivation and design rationale are provided in Appendix~\ref{app:benchmark_motivation}, we propose a three-dimensional quantitative metric that forms a composite quality score $Q$ for an RLVR dataset $\mathcal{D}$, which can be used to predict RLVR training effectiveness. Full coefficient values are listed in Appendix~\ref{app:hyperparams}.

\paragraph{$S_1$: Static Data Quality.}
$S_1$ aggregates three static sub-scores: the verifiability score $S_{1a}$, the learnability score $S_{1b}$, and the contamination robustness score $S_{1c}$.
$S_{1a}$ rewards datasets whose answers are unambiguous and consistently aligned across sources, while penalizing MCQ-heavy data and problems that appear excessively across datasets:
\begin{equation}
\small
    S_{1a}=\alpha_{\mathrm{con}} R_{\mathrm{con}} + \alpha_{\mathrm{mcq}} (1-\beta_{\mathrm{mcq}}R_{\mathrm{mcq}}) + \alpha_{\mathrm{reuse}} P_{\mathrm{reuse}},
\end{equation}
where $R_\mathrm{con}$ denotes the fraction of prompts with consistent answers across all occurrences, $R_\mathrm{mcq}$ is the fraction of MCQs in the dataset, and $P_\mathrm{reuse}$ is a discount factor that penalizes excessively reused problems across datasets.

$S_{1b}$ is derived from the SCA-annotated learnability score $L_\mathrm{SCA}$
and augmented with a diversity bonus $\varepsilon_\mathrm{div}$ based on normalized entropy:
\begin{equation}
\small
    S_{1b} \;=\; \sigma(L_\mathrm{SCA}) + \varepsilon_\mathrm{div},
\label{eq:S1b}
\end{equation}
where $\sigma$ is the sigmoid function used to stabilize the score scale.

$S_{1c}$ measures the proportion of samples in the dataset that are free from confirmed evaluation-set leakage, including both exact-match instances and semantically overlapping instances with evaluation benchmarks.
$S_{1c}$ is defined as:
\begin{equation}
\small
    S_{1c}
    \;=\;
    1 - \frac{N_\mathrm{leak}}{N},
\end{equation}
where $N_\mathrm{leak}$ denotes the number of leaked samples identified through exact or near-duplicate matching, and $N$ is the total number of samples in the dataset.

$S_1$ is then computed as a weighted combination:
\begin{equation}
\small
    S_1 \;=\; w_{1a}\cdot S_{1a} \;+\; w_{1b}\cdot S_{1b} \;+\; w_{1c}\cdot S_{1c}.
\end{equation}

\paragraph{$S_2$: Sampling Efficiency Gain.}
$S_2$ measures improvements in sampling efficiency, reflected by $\mathrm{Mean}@N$. Since empirical RLVR gains may exhibit different levels of reliability across model scales, we adopt a \emph{scale-adaptive} formulation that combines empirical performance gains with SCA-based learnability attribution $L_{\mathrm{SCA}}$:
\begin{equation}
\small
    S_2(\mathcal{D};\,M) \;=\;
        \lambda(M)\cdot \widetilde{\Delta\mathrm{Mean@4}}
        \;+\;
        \bigl(1-\lambda(M)\bigr)\cdot L_{\mathrm{SCA}},
\label{eq:S2}
\end{equation}
where $\widetilde{\Delta\mathrm{Mean@4}}$ denotes the normalized $\mathrm{Mean}@4$ improvement on Math500, which serves as both the validation benchmark and a proxy for RLVR learning effectiveness, and $\lambda(M)$ is a scale-dependent interpolation weight determined by the model scale $M$.

\paragraph{$S_3$: Capability Boundary Expansion.}
$S_3$ measures improvements in $\mathrm{Pass}@N$, thereby capturing capability boundary expansion. Similar to $S_2$, we combine empirical gains with $L_{\mathrm{SCA}}$:
\begin{equation}
\small
    S_3(\mathcal{D};\,M) \;=\;
        \lambda(M)\cdot \widetilde{\Delta\mathrm{Pass@4}}
        \;+\;
        \bigl(1-\lambda(M)\bigr)\cdot L_{\mathrm{SCA}},
\label{eq:S3}
\end{equation}
where $\widetilde{\Delta\mathrm{Pass@4}}$ denotes the normalized $\mathrm{Pass}@4$ improvement on Math500.

\paragraph{Composite RLVR Dataset Quality Score.}
The three dimensions are combined into a single score:
\begin{equation}
\small
    Q(\mathcal{D};\, M) \;=\; w_1(M)\cdot S_1 \;+\; w_2(M)\cdot S_2 \;+\; w_3(M)\cdot S_3.
\label{eq:Q}
\end{equation}
All scale-dependent weights and coefficients follow
$f(M) = a + b\cdot\log_{10}(M/\mathrm{1B})$, 
allowing the scoring preference to transition smoothly across model scales without discrete thresholds.


Intuitively, larger models place greater emphasis on dynamic utility (e.g., capability boundary expansion and sampling efficiency), while smaller models rely more heavily on static data quality and learnability signals.

%% file: section/4_data_curation.tex

%% file: section/5_experiments.tex
\section{Experiments}

\subsection{Experimental Setup}
\paragraph{Dataset.}
Based on the results in Section~\S\ref{sec:data_landscape}, we select DeepScaleR~\cite{deepscaler2025}, OpenR1-Math-220k~\cite{openr1_math_220k}, DeepMath-103k~\cite{he2025deepmath103k}, DAPO-Math-17k~\cite{yu2025dapo}, and Skywork-OR1-RL-Data~\cite{he2025skyworkopenreasoner1} for evaluation, as these datasets exhibit particularly representative and distinctive compositional patterns. Additionally, we construct a modified variant named DAPO++, in which leaked instances identified in the original DAPO are removed and replaced with randomly selected non-MCQ and SCA-annotated learnable instances from the \textit{stack exchange} atomic source uncovered by ATLAS, which were rarely inherited by subsequent RLVR datasets.

\paragraph{Implementation Details.}
The complete set of hyperparameters used in our training and details of the evaluation are provided in Appendix \ref{app:implementation_details}. 
We use \textit{Qwen3-1.7B-Base} and \textit{Qwen3-8B-Base} \cite{yang2025qwen3technicalreport} as the base models and employ GRPO~\cite{grpo2024deepseek} as our RL algorithm. 
We evaluate on eight widely used mathematical reasoning benchmarks: \textbf{AMC23}~\cite{amc23}, \textbf{AIME24}~\cite{aime24}, \textbf{AIME25}~\cite{aime25}, \textbf{AMO}~\cite{an2025amobench}, \textbf{Minerva}~\cite{lewkowycz2022minerva}, \textbf{Olympiad}~\cite{he2024olympiadbench}, \textbf{HLE}~\cite{HLE2026}, and \textbf{MATH-500} \cite{hendrycks2021math}\footnote{Since \textit{Math500} is used as the validation set for checkpoint selection in all experiments, we exclude it from the average score computation and denote the resulting metric as average$^*$.}.
To assess generalization beyond math reasoning, we further test on general reasoning benchmarks \textbf{GPQA-Diamond}~\cite{rein2023gpqa}, with multiple-choice options shuffled to prevent contamination. For the AMC23, AIME24, AIME25, and AMO benchmark, which have relatively small test sets, we report \textbf{\textit{Mean@32}} (for \textit{\textbf{Pass@1}}), while for the other benchmarks we report \textbf{\textit{Mean@4}}.

\paragraph{Reference Dataset Ranking.}

Since the relative performance of RLVR datasets varies across model scales, there is no natural ground-truth ranking for dataset quality. To provide a unified reference for evaluating our proposed benchmarking metrics, we construct a reference dataset ranking using \textit{\textbf{S}td-weighted \textbf{Rank} aggregation} (SRank).
where rankings from model scales with larger cross-dataset performance variance receive higher weights due to their stronger discriminative power.
To account for the discriminative strength of each model scale, we weight each ranking by the cross-dataset standard deviation under that scale:
\begin{equation}
\small
    w_m=\frac{\sigma_m}{\sum_{m'}\sigma_{m'}} ,
\end{equation}
where $\sigma_m$ is the standard deviation of Average$^*$ scores across datasets for model scale $m$. The final aggregated rank score for dataset $D$ is defined as:
\begin{equation}
\small
    SRank(D)=\sum_m w_m \cdot \mathrm{rank}_m(D).
\end{equation}




\subsection{Experimental Results}
\label{sec:result}
\paragraph{Reasoning Performance.}
\input{table/main_results}
\input{table/ref_rank}

\input{table/quailty_score}

\input{table/filter_mcq_mean}
\input{table/decon_exp}

Based on the leakage analysis in Section~\S\ref{sec:data_landscape}, we construct decontaminated versions of five selected representative datasets, train corresponding RLVR checkpoints on them, and evaluate their performance on both mathematical reasoning benchmarks (Table~\ref{tab:main_result}) and general reasoning tasks (Table~\ref{tab:gpqa_result}). 

We first observe substantial differences in dataset discriminability across model scales. Specifically, the performance gaps among datasets are relatively small under the 1.7B setting (std.=0.59), whereas the 8B setting exhibits much clearer dataset separation (std.=2.17). Additionally, while DAPO++ consistently achieves the best overall performance across both model scales, the second best-performing dataset is not consistent across scales: DeepMath-103K performs stronger under the 1.7B setting, whereas DAPO performs better among the remaining baselines under the 8B setting. Therefore, based on our proposed SRank metric (Table~\ref{tab:srank_ref}), we obtain the unified reference ranking of RLVR datasets as follows: DAPO++ (ours) $>$ DAPO $>$ DeepScaleR $>$ DeepMath $>$ Skywork $>$ OpenR1.

\paragraph{Benchmark Result.}

Using the quality scoring framework proposed in Section~\S\ref{sec:rlvr_score}, along with the corresponding SCA-based \textit{Mean@N} and \textit{Pass@N} results reported in Appendix~\ref{app:sca_results}, we obtain the quality scores shown in Table~\ref{tab:quailty_score}. 
First, our quality scores exhibit strong correlation with the averaged mathematical reasoning performance. In particular, the Spearman correlations between the Q scores and Average$^{*}$ performance reach 0.60 and 0.94 under the 1.7B and 8B settings, respectively, indicating that our scoring framework effectively captures dataset utility trends. 
Moreover, compared with the reference ranking constructed by SRank, our scoring method remains highly consistent with the more discriminative 8B ranking (\(\rho=0.93\)), while moderately smoothing the noisier 1.7B ranking (\(\rho=0.54\)). These results collectively demonstrate the validity and robustness of our proposed scoring framework.

\paragraph{Ablation Study.}

We conduct ablation studies on the two key factors of high-quality RLVR data proposed in Static Quality, namely \textit{Verifiability} and \textit{Contamination}.

As shown in Table~\ref{tab:ablation_mcq}, converting MCQ data into open-ended questions consistently improves the Average$^*$ score. This suggests that, compared to multiple-choice formats, open-ended questions provide more reliable and consistent supervision signals, thereby improving reward verifiability and learnability during RLVR training. The results validate the importance of high \textit{Verifiability} in constructing effective RLVR datasets.

Furthermore, as shown in Table~\ref{tab:ablation_decontamination}, removing contaminated data does not hurt performance; instead, it leads to overall improvements, with the largest gains observed on challenging benchmarks such as AIME. This indicates that leaked data are not necessarily useful difficult data, but may instead contain high-noise or low-learnability signals. Moreover, performance improvements obtained by existing methods on contaminated datasets do not necessarily imply genuine reasoning capability, but may partially stem from memorizing leaked evaluation instances themselves. These findings further underscore the significance of our systematic data collection and lineage tracing framework, which makes large-scale global decontamination across RLVR corpora possible, thereby enabling the construction of cleaner, more reliable, and contamination-controlled training datasets.

%% file: table/main_results.tex
\begin{table*}[t]
\centering
\small
\setlength{\tabcolsep}{2.5pt}
\renewcommand{\arraystretch}{1.08}
\begin{tabular}{lcccccccc|c}
\toprule
& \multicolumn{4}{c}{\textbf{Mean@4}} & \multicolumn{4}{c}{\textbf{Mean@32}} & \\
\cmidrule(lr){2-5}\cmidrule(lr){6-9}
\textbf{Source} & \textbf{Math500$^*$} & \textbf{Minerva} & \textbf{Olympiad} & \textbf{HLE} & \textbf{AMC23} & \textbf{AIME24} & \textbf{AIME25} & \textbf{AMO} & \textbf{Average$^*$} \\
\midrule

\textbf{\textit{Qwen3-1.7B-Base}}
& 48.4 & 15.3 & 17.6 & 5.9 & 28.2 & 4.7 & 2.3 & 1.3 & 10.8 \\

$\hookrightarrow$ DeepScaleR
& 58.1 & 23.3 & 23.7 & 6.3 & 34.5 & 6.3 & 7.6 & 1.0 & 14.7 \\

$\hookrightarrow$ DeepMath-103K
& 57.9 & 26.5 & 23.1 & 8.4 & 35.4 & 8.4 & 3.1 & 3.2 & \underline{15.4} \\

$\hookrightarrow$ OpenR1-Math-220k
& 58.9 & 23.1 & 24.0 & 4.7 & 31.6 & 6.9 & 6.3 & 1.4 & 14.0 \\

$\hookrightarrow$ DAPO-Math-17k
& 58.5 & 22.6 & 22.9 & 5.9 & 39.5 & 7.0 & 5.9 & 1.3 & 15.0 \\

$\hookrightarrow$ Skywork-OR1-RL-Data
& 58.8 & 22.8 & 24.7 & 5.1 & 40.9 & 5.8 & 4.6 & 2.0 & 15.1 \\

\rowcolor{tblue} $\hookrightarrow$ DAPO++ (ours)
& 60.0 & 21.9 & 24.9 & 6.8 & 41.9 & 9.4 & 3.3 & 1.7 & \textbf{15.7} \\

\midrule
\addlinespace[2pt]
\textbf{\textit{Qwen3-8B-Base}}
& 61.6 & 25.6 & 27.9 & 5.7 & 48.6 & 10.4 & 10.5 & 1.8 & 18.6 \\

$\hookrightarrow$ DeepScaleR
& 75.9 & 34.7 & 38.8 & 4.5 & 61.6 & 23.4 & 17.1 & 2.9 & 26.1 \\

$\hookrightarrow$ DeepMath-103K
& 73.4 & 31.9 & 39.2 & 4.5 & 62.7 & 17.1 & 16.7 & 3.7 & 25.1 \\

$\hookrightarrow$ OpenR1-Math-220k
& 73.1 & 35.7 & 37.0 & 3.9 & 63.2 & 16.5 & 15.4 & 3.3 & 25.0 \\

$\hookrightarrow$ DAPO-Math-17k
& 77.8 & 36.7 & 43.0 & 4.1 & 69.8 & 27.9 & 21.5 & 2.4 & \underline{29.3} \\

$\hookrightarrow$ Skywork-OR1-RL-Data
& 73.2 & 34.5 & 37.8 & 3.1 & 64.1 & 17.3 & 15.8 & 2.8 & 25.1 \\

\rowcolor{tblue} $\hookrightarrow$ DAPO++ (ours)
& 77.7 & 37.1 & 43.5 & 4.7 & 69.3 & 25.9 & 22.6 & 3.9 & \textbf{29.6} \\

\bottomrule
\end{tabular}
\caption{RLVR results on the \textbf{\textit{decontaminated versions of datasets}} after removing leaked or test-set-related instances.}
\label{tab:main_result}
\end{table*}

%% file: table/ref_rank.tex
\begin{table}[htbp]
\centering
\small

\begin{tabular}{lccc}
\toprule
\multirow{2}{*}{\textbf{Dataset}} & \textbf{Qwen3-1.7B} & \textbf{Qwen3-8B} & \textbf{SRank}$\downarrow$ \\
 & \textbf{\scriptsize Avg. (Rank)} & \textbf{\scriptsize Avg. (Rank)} & \textbf{\scriptsize Score (Rank)} \\
\midrule
DAPO++      & 15.7 (1) & 29.6 (1) & 1.00 (1) \\
DAPO        & 15.0 (4) & 29.3 (2) & 2.43 (2) \\
DeepScaleR  & 14.7 (5) & 26.1 (3) & 3.43 (3) \\
DeepMath    & 15.4 (2) & 25.1 (4) & 3.57 (4) \\
Skywork     & 15.1 (3) & 25.1 (4) & 3.79 (5) \\
OpenR1      & 14.0 (6) & 25.0 (6) & 6.00 (6) \\
\bottomrule
\end{tabular}

\caption{Reference dataset ranking constructed using SRank. The final score of SRank places DAPO as the top-ranked RLVR dataset.}
\label{tab:srank_ref}
\end{table}

%% file: table/quailty_score.tex
\begin{table}[t]
\small
\centering
\begin{threeparttable}
\setlength{\tabcolsep}{5pt}
\begin{tabular}{lcccc}
\toprule
\textbf{Dataset} & \textbf{S1} & \textbf{S2} & \textbf{S3} & \textbf{Q$\uparrow$ (Rank)} \\
\midrule

\multicolumn{5}{c}{\textit{\textbf{Qwen3-1.7B-Base}}} \\
\midrule
DeepScaleR      & 0.901 & 0.685 & 0.616 & 0.804 (5) \\
DeepMath        & 0.980 & 0.795 & 0.720 & \underline{0.895} (2) \\
OpenR1          & 0.906 & 0.511 & 0.443 & 0.738 (6) \\
DAPO            & 0.969 & 0.836 & 0.788 & \textbf{0.909} (1) \\
Skywork         & 0.944 & 0.796 & 0.740 & 0.876 (4) \\
\rowcolor{tblue} DAPO++ (ours) & 0.967 & 0.728 & 0.772 & 0.878 (3)  \\

\midrule
Pearson $r$     & +0.81 & +0.75 & +0.89 & +0.85 \\
Spearman $\rho$ & +0.66 & +0.43 & +0.60 & +0.60 \\

\midrule
\multicolumn{5}{c}{\textit{\textbf{Qwen3-8B-Base}}} \\
\midrule
DeepScaleR      & 0.811 & 0.678 & 0.442 & 0.654 (2) \\
DeepMath        & 0.890 & 0.569 & 0.289 & 0.598 (3) \\
OpenR1          & 0.825 & 0.539 & 0.212 & 0.541 (5) \\
DAPO            & 0.897 & 0.772 & 0.539 & \underline{0.746} (2) \\
Skywork         & 0.857 & 0.561 & 0.206 & 0.558 (4) \\
\rowcolor{tblue} DAPO++ (ours) & 0.929 & 0.778 & 0.524 & \textbf{0.754} (1) \\

\midrule
Pearson $r$     & +0.70 & +0.96 & +0.91 & +0.96 \\
Spearman $\rho$ & +0.60 & +0.94 & +0.77 & +0.94 \\
\bottomrule
\end{tabular}

\vspace{0.25em}
{\scriptsize\raggedright
$S_1$: \textit{Static Data Quality}, 
$S_2$: \textit{Sampling Efficiency}, 
$S_3$: \textit{Capability Boundary Expansion}, and
$Q$: \textit{Overall Quailty Score}. \par}

\caption{Multi-dimensional quality scores. Pearson $r$ and Spearman $\rho$ are computed between the Q scores and Average$^{*}$ results in Table~\ref{tab:main_result}.}
\label{tab:quailty_score}
\end{threeparttable}
\end{table}

%% file: table/filter_mcq_mean.tex










\begin{table*}[t]
\centering
\small
\setlength{\tabcolsep}{4.5pt}
\renewcommand{\arraystretch}{1.08}
\begin{tabular}{lcccccccc|c}
\toprule
\textbf{Source} & \textbf{Math500$^*$} & \textbf{Minerva} & \textbf{Olympiad} & \textbf{HLE} & \textbf{AMC23} & \textbf{AIME24} & \textbf{AIME25} & \textbf{AMO} & \textbf{Average$^*$} \\
\midrule

\textbf{\textit{Qwen3-1.7B-Base}} & 48.4 & 15.3 & 17.6 & 5.9 & 28.2 & 4.7 & 2.3 & 1.3 & 10.8 \\
\midrule

olympiads
& 57.8 & 22.3 & 23.1 & 6.1 & 35.1 & 5.3 & 5.2 & 2.1 & 14.2 \\
\rowcolor{tblue} $\hookrightarrow$ w/o mcq
& 59.0 & 24.1 & 23.2 & 5.5 & 36.2 & 9.4 & 3.3 & 1.8 & \textbf{14.8} \\



\addlinespace[2pt]

amc\_aime
& 58.7 & 23.0 & 23.6 & 7.0 & 36.8 & 8.4 & 4.3 & 1.9 & 15.0 \\
\rowcolor{tblue} $\hookrightarrow$ w/o mcq
& 59.1 & 21.5 & 23.3 & 5.5 & 46.0 & 10.4 & 5.7 & 2.5 & \textbf{16.4} \\

\bottomrule
\end{tabular}
\caption{Mean@N results for MCQ ablation.}
\label{tab:ablation_mcq}
\end{table*}

%% file: table/decon_exp.tex
\begin{table*}[t]
\centering
\small
\setlength{\tabcolsep}{2.5pt}
\renewcommand{\arraystretch}{1.08}
\begin{tabular}{llcccccccc|c}
\toprule
\textbf{Model} & \textbf{Dataset} & \textbf{Math500$^*$} & \textbf{Minerva} & \textbf{Olympiad} & \textbf{HLE} & \textbf{AMC23} & \textbf{AIME24} & \textbf{AIME25} & \textbf{AMO} & \textbf{Average$^*$} \\
\midrule


& DeepScaleR
& 58.7 & 22.3 & 23.9 & 5.5 & 33.7 & 6.3 & 4.3 & 2.8 & 14.1 \\

\textbf{\textit{Qwen3-1.7B}}
& \cellcolor{tblue}$\hookrightarrow$ Decontam.
& \cellcolor{tblue}58.1 & \cellcolor{tblue}23.3 & \cellcolor{tblue}23.7 & \cellcolor{tblue}6.3
& \cellcolor{tblue}34.5 & \cellcolor{tblue}6.3 & \cellcolor{tblue}7.6 & \cellcolor{tblue}1.0 & \cellcolor{tblue}\textbf{14.7} \\

\addlinespace[2pt]

\textbf{\textit{Base}}
& DAPO-Math-17k
& 59.7 & 22.2 & 24.0 & 6.8 & 37.1 & 5.3 & 5.0 & 1.9 & 14.6 \\
& \cellcolor{tblue}$\hookrightarrow$ Decontam.
& \cellcolor{tblue}58.5 & \cellcolor{tblue}22.6 & \cellcolor{tblue}22.9 & \cellcolor{tblue}5.9 & \cellcolor{tblue}39.5 & \cellcolor{tblue}7.0 & \cellcolor{tblue}5.9 & \cellcolor{tblue}1.3 & \cellcolor{tblue}\textbf{15.0} \\ 

\midrule

& DeepScaleR
& 73.0 & 33.9 & 37.0 & 3.5 & 62.7 & 16.9 & 14.2 & 2.8 & 24.4 \\

\textbf{\textit{Qwen3-8B}}
& \cellcolor{tblue}$\hookrightarrow$ Decontam.
& \cellcolor{tblue}75.9 & \cellcolor{tblue}34.7 & \cellcolor{tblue}38.8 & \cellcolor{tblue}4.5
& \cellcolor{tblue}61.6 & \cellcolor{tblue}23.4 & \cellcolor{tblue}17.1 & \cellcolor{tblue}2.9 & \cellcolor{tblue}\textbf{26.1} \\ 

\addlinespace[2pt]

\textbf{\textit{Base}}
& DAPO-Math-17k
& 79.4 & 36.8 & 43.6 & 4.1 & 68.8 & 28.1 & 21.7 & 2.6 & 29.4 \\
& \cellcolor{tblue}$\hookrightarrow$ Decontam.
& \cellcolor{tblue}77.7 & \cellcolor{tblue}35.7 & \cellcolor{tblue}43.7 & \cellcolor{tblue}4.3 & \cellcolor{tblue}69.6 & \cellcolor{tblue}28.4 & \cellcolor{tblue}22.7 & \cellcolor{tblue}2.8 & \cellcolor{tblue}\textbf{29.6} \\

\bottomrule
\end{tabular}
\caption{Mean@N results for decontamination ablation across mathematical reasoning benchmarks.}
\label{tab:ablation_decontamination}
\end{table*}

%% file: section/6_conclusion.tex
\section{Conclusion}

In this work, we propose \textit{ATLAS}, a lineage-aware atomic-source tracing framework for large-scale provenance analysis in RLVR datasets. Using ATLAS, we uncover the hidden composition of widely used RLVR datasets, revealing substantial overlap in their upstream sources and significant contamination risks. Guided by these findings, we further curate a new RLVR dataset, \textsc{DAPO++}. Beyond provenance tracing, we introduce SCA, a source-aware attribution framework for RLVR dataset benchmarking and quality estimation, together with a composite dataset quality score $Q$ for predicting downstream RLVR effectiveness. Overall, our work provides a roadmap for RLVR dataset construction, benchmarking, and selection, laying the foundation for future RLVR research.

%% file: section/limitations.tex
\section*{Limitations}
Although ATLAS traces over one million RLVR instances back to their atomic sources, the tracing process still relies heavily on manual curation and human audit. In particular, many early mathematical reasoning datasets exhibit severe nested reuse and ambiguous provenance chains, requiring careful inspection of original papers and dataset documentation to disentangle hidden dependencies (e.g., \texttt{draw}, and \texttt{asdiv}). Due to the substantial manual effort required for reliable attribution, as well as the incomplete availability and gradual loss of older papers and dataset resources, our current tracing only recovers several lineage branches back to approximately 2014-era datasets (e.g., \texttt{AddSub} and \texttt{SimulEq}). Therefore, older data sources remain to be incorporated in future work.

%% file: section/appendix.tex
\section{Details of Data Lineage Tracing}

\subsection{Collected RLVR Datasets and Traced Atomic Sources}
\label{sec:detail_data_collection}
\input{table/data_info}

\input{table/source_info}
Based on the limited data provenance described in the HuggingFace repositories, blogs or associated papers of these RLVR datasets, we conduct an iterative and labor-intensive data lineage analysis to identify 5 core data sources. Among them, 3 are large-scale aggregated data that compile substantial amounts of prior datasets, while the remaining 2 are more atomic sources, a non-trivial portion of whose contents has been reused in subsequent datasets without explicit attribution.
Detailed statistics of all datasets, including their HuggingFace ids, corresponding papers, and dataset sizes, are provided in Table \ref{tab:dataset_details}. 
Meanwhile, Table~\ref{tab:data_source_detail} summarizes the atomic sources traced by ATLAS. The problem-type labels are manually annotated through case inspection and careful examination of the associated papers and dataset construction documentation.

\subsection{ATLAS Pipeline}
\label{app:data_lineage_tracing}
\input{table/algo}
ATLAS is composed of three stages, and the overall tracing procedure is summarized in Algorithm~\ref{alg:lineage_tracing}.

\subsection{RLVR Dataset Provenance}
\label{app:rlvr_provenance}

\begin{figure*}[htbp]
\centering
  \includegraphics[width=\linewidth]{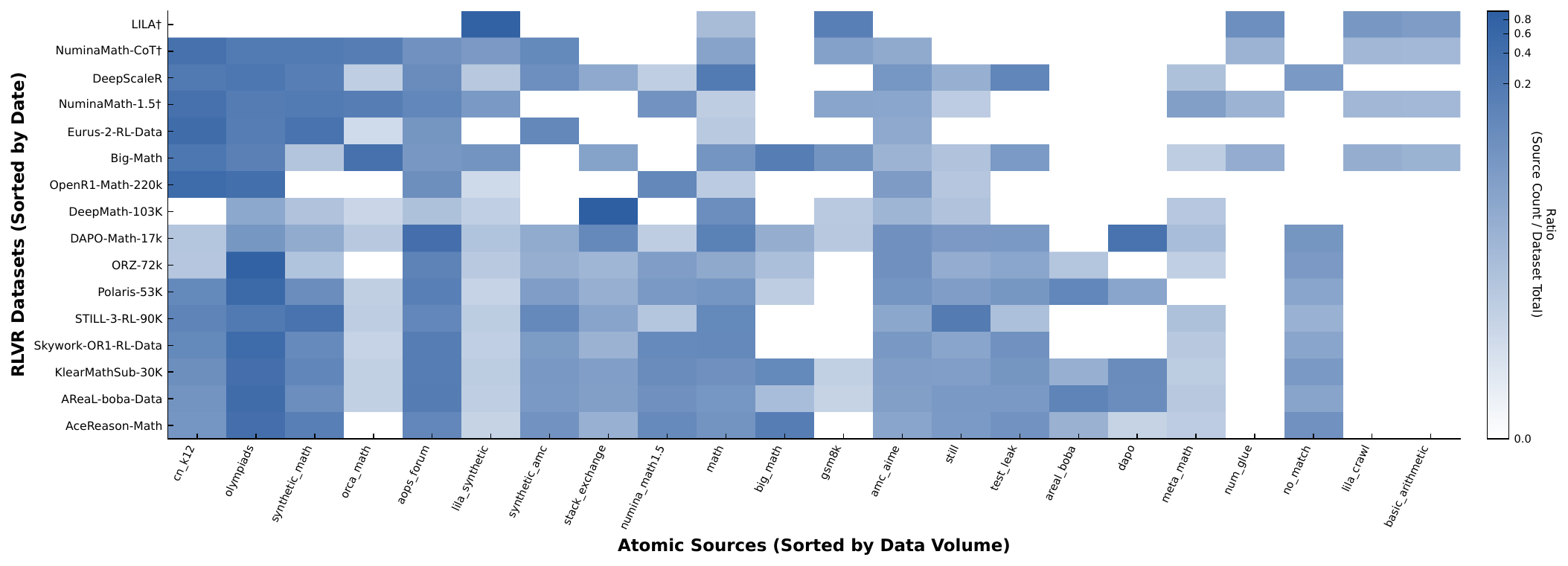}
  \caption{Atomic-source distributions across datasets. Reading each column from top to bottom, the first colored cell denotes the dataset in which the corresponding atomic source was first introduced into the RLVR data ecosystem.}
  \label{fig:source_heatmap}
\end{figure*}

\begin{figure*}[htbp]
    \centering
    \includegraphics[width=\linewidth]{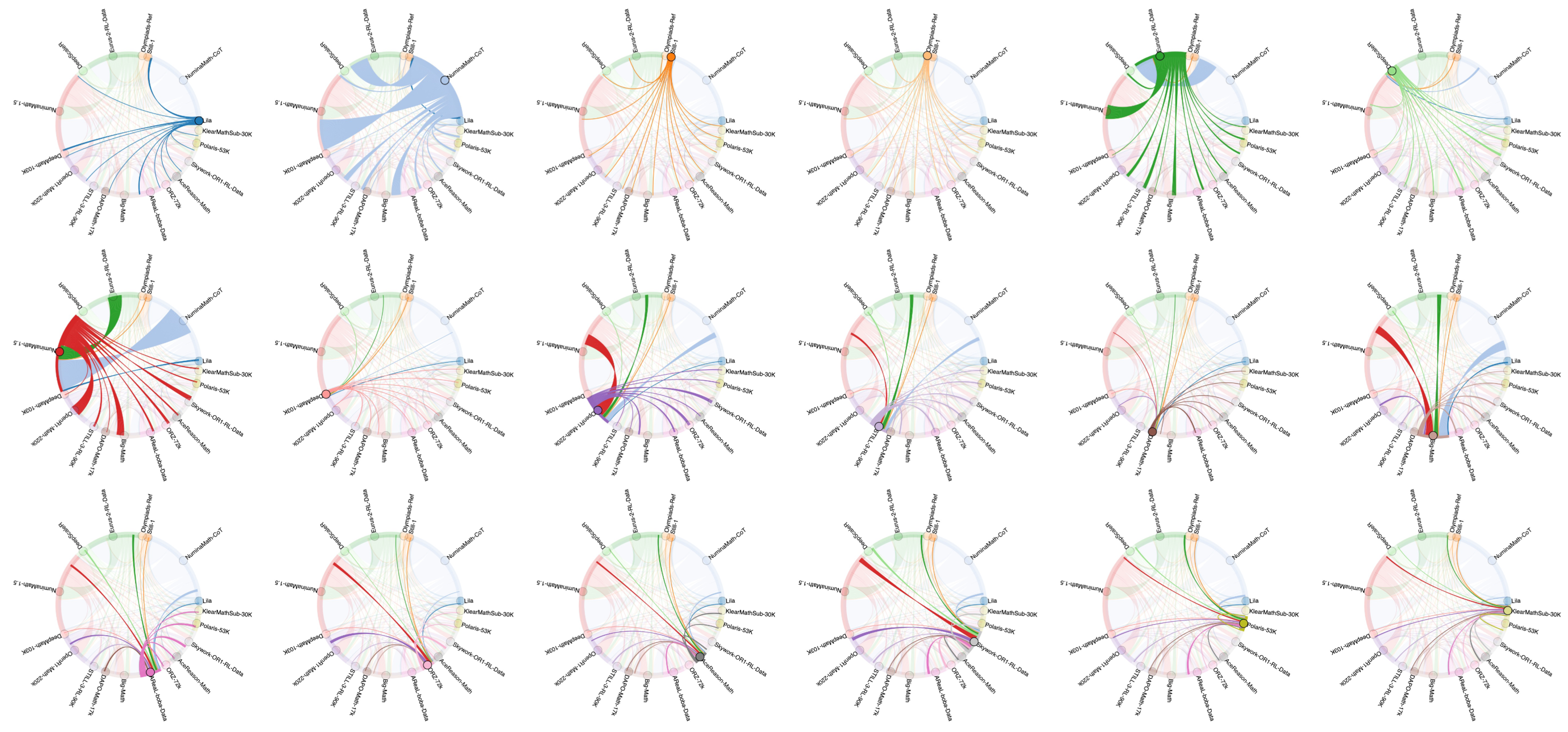}
    \caption{Bokeh plots visualizing the compositional relationships and data overlaps among different RLVR datasets.}
    \label{fig:bokeh_grid_split}
\end{figure*}

Figure~\ref{fig:source_heatmap} further reveals the specific RLVR datasets in which these atomic data sources were first introduced. Although many datasets exhibit substantial atomic source overlap,
particularly with the NuminaMath-CoT and NuminaMath-1.5 series~\cite{numina_math_datasets}, we still observe considerable heterogeneity in the relative source proportions across datasets. 
We further visualize the compositional relationships among RLVR datasets using Bokeh plots, where the detailed views for each dataset are provided in Figure~\ref{fig:bokeh_grid_split}.

\section{Details of Leakage Analysis}

\subsection{Source Leakage}
\label{app:leakage_analysis}

\begin{figure}[htbp]
\centering
  \includegraphics[width=\linewidth]{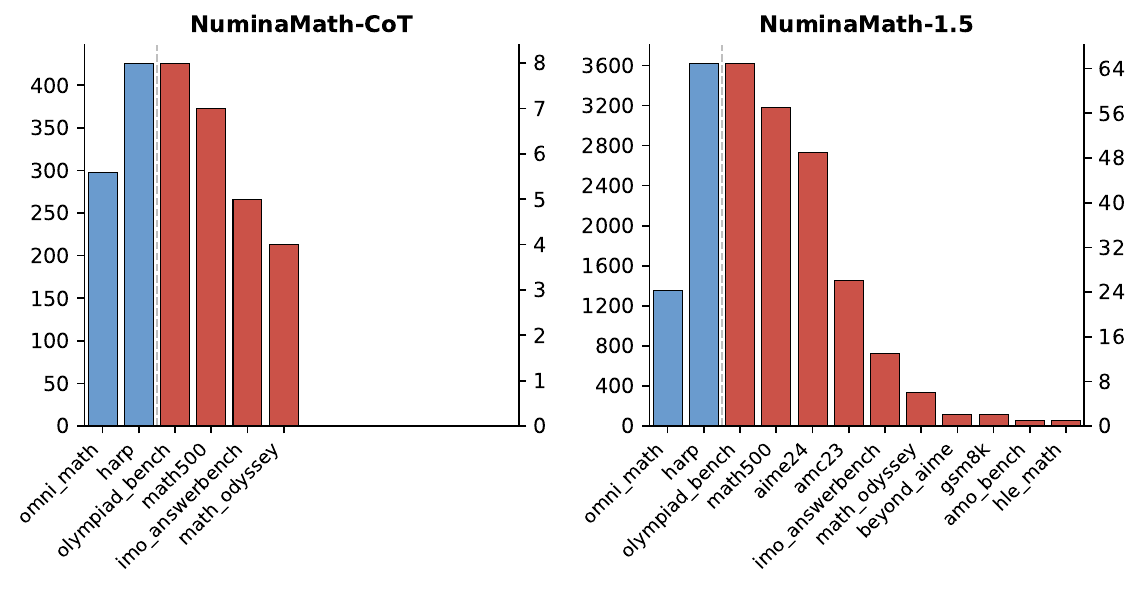}
  \caption{Benchmark leakage statistics of main source datasets, measured by semantic overlap.}
  \label{fig:leak_per_source_dataset}
\end{figure}

Figure~\ref{fig:leak_per_source_dataset} presents the benchmark leakage statistics of major source datasets measured by semantic overlap. We primarily report high-risk leakage cases with semantic similarity scores exceeding 90\%, while noting that a small number of potentially leaked cases may remain in the 80\% similarity range.

\subsection{Leakage Case Studies}
\label{app:leakage_analysis_case}
\input{table/case_study}

Table~\ref{tab:case_study} presents representative case studies of benchmark leakage identified through our lineage-aware tracing and semantic matching pipeline.

\section{Details of RLVR Dataset Benchmarking}
\subsection{What Makes Good RLVR Data?}
\label{app:benchmark_motivation}
We evaluate RLVR data from two complementary perspectives: \textbf{Static Quality} and \textbf{Dynamic Utility}.

\paragraph{Static Quality.}
Static Quality measures the intrinsic properties of an instance and whether it provides reliable and informative supervision signals. We consider three criteria:
(1) \textbf{Verifiability}: Answers should be unambiguous and automatically verifiable, such that identical prompts should not correspond to conflicting answers. Open-ended formats are generally preferred over multiple-choice questions (MCQs), whose answers may be guessed without genuine reasoning.
(2) \textbf{Learnability}: samples should provide meaningful learning signals, avoiding problems that are overly easy or overly difficult.
(3) \textbf{Low Contamination}: training data should have minimal overlap with evaluation benchmarks, including both exact duplicates and semantically similar problems.

\paragraph{Dynamic Utility.}
Dynamic Utility evaluates whether the instance from a particular atomic source leads to actual performance gains after RLVR training while independent of test set. Assessing with a held-out benchmark, we consider two metrics:
(1) \textbf{Sampling Efficiency Gain}: the improvement in sampling efficiency of the RL-aligned model relative to the base model, measured by the increase in $\mathrm{Mean}@N$.
(2) \textbf{Capability Boundary Expansion}: the extent to which RL training enables the model to solve previously unsolved problems relative to the base model, measured by improvements in $\mathrm{Pass}@N$.

\subsection{Details of Implementation}
\label{app:implementation_details}

\paragraph{Training Settings}
For RL finetuning, we use the widely adopted GRPO algorithm built on \textit{VERL} \cite{2025verlframework} framework. The full hyperparameters used in our training are listed in Table~\ref{tab:hparam}. Specifically, for rollout generation, we use a temperature of 1.0, and rewards are computed using \textit{Math-Verify}. All models are trained for $T=500$ optimization steps, and we report results using \textit{Math500} as validation set to select best checkpoint.
\input{table/hyperparameters}

\paragraph{Evaluation Settings}
For all evaluations, we use a decoding temperature of 0.6 with \texttt{top\_p}=0.95, \texttt{top\_k}=20, a maximum generation length of 16384 tokens, and a fixed random seed of 0. For the AMC23, AIME24, AIME25, and AMO benchmarks, which contain relatively small test sets, we report \textbf{\textit{Mean@32}} (for \textit{\textbf{Pass@1}}), while for the remaining benchmarks we report \textbf{\textit{Mean@4}} (for \textit{\textbf{Pass@1}}).

\subsection{OOD Results}
In addition to mathematical reasoning benchmarks, we also evaluate our method on out-of-domain (OOD) datasets, including the general QA benchmark GPQA, with results shown in Table~\ref{tab:gpqa_result}.

\input{table/gpqa_results}

\subsection{SCA Results}
\label{app:sca_results}

\input{table/source_meann_color}
\input{table/source_passn_color}

We report the evaluation \textit{Mean@N} results (Table~\ref{tab:mean_theta_vs_base}) and \textit{Pass@N} results (Table~\ref{tab:pass_theta_vs_base}) of RLVR checkpoints trained on individual atomic sources identified by ATLAS. Following prior RLVR evaluation practice, we use \textit{Math500} as the validation set and treat its performance as a proxy signal for our dataset scoring framework in Section~\S\ref{sec:rlvr_score}.

\subsection{Hyperparameter Settings}
\label{app:hyperparams}

We provide the hyperparameter settings used for benchmarking in Table~\ref{tab:hyperparams_benchmark}. All hyperparameters were selected through grid search.

\paragraph{Description.}
For the verifiability-related coefficients, $\alpha_{\mathrm{con}}$ controls the weight of the consistency score $R_{\mathrm{con}}$, $\alpha_{\mathrm{mcq}}$ and $\beta_{\mathrm{mcq}}$ jointly determine the strength of the MCQ penalty term, and $\alpha_{\mathrm{reuse}}$ controls the reuse discount factor $P_{\mathrm{reuse}}$. 
For the static quality score $S_1$, the weights $w_{1a}$, $w_{1b}$, and $w_{1c}$ correspond to the contributions of verifiability, learnability, and contamination, respectively. 
The $\alpha$ coefficients are shared across $L_\mathrm{SCA}$, the learnability term $S_{1b}$, and the proxy objectives for $S_2$ and $S_3$. Specifically, $\alpha_{01}$ represents the gold-signal reward, $\alpha_{11}$ corresponds to the too-easy penalty, $\alpha_{10}$ models the base-competence signal, and $\alpha_{00}$ controls the too-hard penalty. 
Finally, the composite quality score $Q$ is formed by aggregating static quality ($S_1$), sampling efficiency ($S_2$), and capability expansion ($S_3$) using weights $w_1$, $w_2$, and $w_3$, respectively.

\input{table/hyperparameters_benchmark}

\section{Related Work}
\label{app:related_work}
Recent surveys have highlighted the growing importance of data provenance, traceability, and transparency in LLMs~\cite{hohensinner2026tracingdatatrailsurvey}, emphasizing the increasing opacity of modern training corpora and the need for provenance-aware AI systems. However, existing work has rarely investigated in the RLVR setting, particularly at the level of \textit{atomic sources}. Prior studies have instead focused primarily on data contamination tracing and contamination detection in LLMs~\cite{cheng2025surveydatacontaminationlarge,fu-etal-2025-ContaminationDetection}.

Moreover, our work also studies RLVR datasets from a benchmarking perspective. While dataset benchmarking has recently attracted increasing attention, existing efforts mainly focus on evaluating \textit{test benchmarks} rather than \textit{training datasets}. For example, Benchmark$^2$~\cite{qian2026benchmark2systematicevaluationllm} systematically evaluates the quality of LLM benchmarks from perspectives such as ranking consistency, discrimination ability, and capability-level alignment, revealing reliability and separability issues in existing evaluation benchmarks. Similarly, Benchmark Health Index ~\cite{zhu2026benchmarkhealthindexsystematic} proposes a data-driven framework for assessing benchmark quality and longevity through discrimination power, anti-saturation capability, and ecosystem impact. In contrast, our work focuses on benchmarking RLVR training datasets through source-aware lineage tracing and counterfactual utility estimation.

%% file: table/data_info.tex
\begin{table*}[htbp]
\small
\setlength{\tabcolsep}{2pt}
\centering
\begin{tabular}{l l l l c@{\hspace{0.5pt}}c}
\toprule
\textbf{Dataset} & \textbf{Date} & \textbf{Huggingface} & \textbf{Length} & 
\multicolumn{2}{l}{\textbf{Link}} \\
[1.5pt]
\midrule
\multicolumn{6}{l}{\textbf{Data Source (Data Source Collection$\dagger$ / Atomic Source)}}\\[3pt]

Lila$\dagger$ \cite{mishra-etal-2022-lila} & 2022.10.11 & \texttt{allenai/lila} & 317k &
\href{https://huggingface.co/datasets/allenai/lila}{\scriptsize\faExternalLink*} & \href{https://aclanthology.org/2022.emnlp-main.392/}{\scriptsize\faIcon{file-alt}} \\

NuminaMath-CoT$\dagger$ \cite{numina_math_datasets} & 2024.07.22 & \texttt{AI-MO/NuminaMath-CoT} & 859k &
\href{https://huggingface.co/datasets/AI-MO/NuminaMath-CoT}{\scriptsize\faExternalLink*} &
\href{https://github.com/project-numina/aimo-progress-prize/blob/main/report/numina_dataset.pdf}{\scriptsize\faIcon{file-alt}} \\

Still-1 \cite{still-1-2024} & 2024.12.12 & \texttt{RUC-AIBOX/long\_form\_thought\_data\_5k} & 4.92k & \href{https://huggingface.co/datasets/RUC-AIBOX/long_form_thought_data_5k}{\scriptsize\faExternalLink*} & \href{https://arxiv.org/abs/2412.09413}{\scriptsize\faIcon{file-alt}} \\

Olympiads-Ref \cite{numina_math_datasets} & 2024.12.20 & \texttt{AI-MO/olympiads-ref-base} & 13.1k & \href{https://huggingface.co/datasets/AI-MO/olympiads-ref-base}{\scriptsize\faExternalLink*} & -- \\

NuminaMath-1.5$\dagger$ \cite{numina_math_datasets} & 2025.02.10 & \texttt{AI-MO/NuminaMath-1.5} & 896k &
\href{https://huggingface.co/datasets/AI-MO/NuminaMath-1.5}{\scriptsize\faExternalLink*} &
-- \\
[1.5pt]
\midrule

\multicolumn{6}{l}{\textbf{RLVR Datasets (Curated with Verifiable Rewards)}}\\[3pt]

Eurus-2-RL-Data \cite{cui2025-Eurus-2-RL-Data} & 2025.02.03 & \texttt{PRIME-RL/Eurus-2-RL-Data} & 481k & \href{https://huggingface.co/datasets/PRIME-RL/Eurus-2-RL-Data}{\scriptsize\faExternalLink*} & \href{https://arxiv.org/abs/2502.01456}{\scriptsize\faIcon{file-alt}} \\

DeepScaleR \cite{deepscaler2025} & 2025.02.10 & \texttt{agentica-org/DeepScaleR-Preview-Dataset} & 40.3k & \href{https://huggingface.co/datasets/agentica-org/DeepScaleR-Preview-Dataset}{\scriptsize\faExternalLink*} & \href{https://pretty-radio-b75.notion.site/DeepScaleR-Surpassing-O1-Preview-with-a-1-5B-Model-by-Scaling-RL-19681902c1468005bed8ca303013a4e2}{\scriptsize\faIcon{file-alt}} \\

DeepMath-103K \cite{he2025deepmath103k} & 2025.02.15 & \texttt{zwhe99/DeepMath-103K} & 103k &
\href{https://huggingface.co/datasets/zwhe99/DeepMath-103K}{\scriptsize\faExternalLink*} &
\href{https://arxiv.org/abs/2504.11456}{\scriptsize\faIcon{file-alt}} \\

OpenR1-Math-220k \cite{openr1_math_220k} & 2025.02.18 & \texttt{open-r1/OpenR1-Math-220k} & 220k & \href{https://huggingface.co/datasets/open-r1/OpenR1-Math-220k}{\scriptsize\faExternalLink*} & -- \\

STILL-3-RL-90K \cite{chen2025still3} & 2025.03.06 & \texttt{RUC-AIBOX/STILL-3-RL-90K} & 88.1k &
\href{https://huggingface.co/datasets/RUC-AIBOX/STILL-3-RL-90K}{\scriptsize\faExternalLink*} &
\href{https://arxiv.org/abs/2503.04548}{\scriptsize\faIcon{file-alt}} \\

DAPO-Math-17k \cite{yu2025dapo} & 2025.03.18 & \texttt{BytedTsinghua-SIA/DAPO-Math-17k} & 17k &
\href{https://huggingface.co/datasets/BytedTsinghua-SIA/DAPO-Math-17k}{\scriptsize\faExternalLink*} &
\href{https://arxiv.org/abs/2503.14476}{\scriptsize\faIcon{file-alt}} \\


Big-Math \cite{albalak2025bigmath} & 2025.03.25 & \texttt{open-r1/Big-Math-RL-Verified-Processed} & 216k &
\href{https://huggingface.co/datasets/open-r1/Big-Math-RL-Verified-Processed}{\scriptsize\faExternalLink*} &
\href{https://arxiv.org/abs/2502.17387}{\scriptsize\faIcon{file-alt}} \\

AReaL-boba-Data \cite{areal_boba} & 2025.03.29 & \texttt{inclusionAI/AReaL-boba-Data} & 106k &
\href{https://huggingface.co/datasets/inclusionAI/AReaL-boba-Data}{\scriptsize\faExternalLink*} &
\href{https://www.inclusion-ai.org/blog/areal-boba/}{\scriptsize\faIcon{file-alt}}  \\

ORZ-72k \cite{hu2025openreasonerzeroopensourceapproach} & 2025.04.06 & \texttt{Open-Reasoner-Zero/orz\_math\_72k\_collection} & 72.4k &
\href{https://huggingface.co/datasets/Open-Reasoner-Zero/orz_math_72k_collection_extended}{\scriptsize\faExternalLink*} &
\href{https://arxiv.org/abs/2503.24290}{\scriptsize\faIcon{file-alt}} \\

AceReason-Math \cite{chen2025acereason-nemotron} & 2025.05.22 & \texttt{nvidia/AceReason-Math} & 49.6k &
\href{https://huggingface.co/datasets/nvidia/AceReason-Math}{\scriptsize\faExternalLink*} &
\href{https://arxiv.org/abs/2505.16400}{\scriptsize\faIcon{file-alt}} \\


Skywork-OR1-RL-Data \cite{he2025skyworkopenreasoner1} & 2025.05.29 & \texttt{Skywork/Skywork-OR1-RL-Data} & 105k &
\href{https://huggingface.co/datasets/Skywork/Skywork-OR1-RL-Data}{\scriptsize\faExternalLink*} &
\href{https://capricious-hydrogen-41c.notion.site/Skywork-Open-Reaonser-Series-1d0bc9ae823a80459b46c149e4f51680}{\scriptsize\faIcon{file-alt}} \\

Polaris-53K \cite{Polaris2025} & 2025.06.18 & \texttt{POLARIS-Project/Polaris-Dataset-53K} & 53.3k &
\href{https://huggingface.co/datasets/POLARIS-Project/Polaris-Dataset-53K}{\scriptsize\faExternalLink*} &
\href{https://hkunlp.github.io/blog/2025/Polaris/}{\scriptsize\faIcon{file-alt}} \\

KlearMathSub-30K \cite{su2026klearreasoner} & 2025.08.11 & \texttt{Kwai-Klear/KlearReasoner-MathSub-30K} & 30k &
\href{https://huggingface.co/datasets/Kwai-Klear/KlearReasoner-MathSub-30K}{\scriptsize\faExternalLink*} &
\href{https://arxiv.org/abs/2508.07629}{\scriptsize\faIcon{file-alt}} \\

\midrule
\multicolumn{6}{l}{\textbf{Mathematical Reasoning Benchmarks (Test Sets)}}\\[3pt]

MATH-500 \cite{hendrycks2021math} & 2021.03.05 & \texttt{HuggingFaceH4/MATH-500} & -- &
\href{https://huggingface.co/datasets/HuggingFaceH4/MATH-500}{\scriptsize\faExternalLink*} 
& \href{https://arxiv.org/abs/2103.03874}{\scriptsize\faIcon{file-alt}} \\

GSM8K \cite{cobbe2021gsm8k} & 2021.10.27 & \texttt{openai/gsm8k} & -- 
& \href{https://huggingface.co/datasets/openai/gsm8k}{\scriptsize\faExternalLink*}
& \href{https://arxiv.org/abs/2110.14168}{\scriptsize\faIcon{file-alt}} \\

MinervaMath \cite{lewkowycz2022minerva} & 2022.06.29 & \texttt{math-ai/minervamath} & -- 
& \href{https://huggingface.co/datasets/math-ai/minervamath}{\scriptsize\faExternalLink*}
& \href{https://proceedings.neurips.cc/paper_files/paper/2022/file/18abbeef8cfe9203fdf9053c9c4fe191-Paper-Conference.pdf}{\scriptsize\faIcon{file-alt}} \\

OlympiadBench \cite{he2024olympiadbench} & 2024.02.21 & \texttt{knoveleng/OlympiadBench} & -- 
& \href{https://huggingface.co/datasets/knoveleng/OlympiadBench}{\scriptsize\faExternalLink*}
& \href{https://arxiv.org/abs/2402.14008}{\scriptsize\faIcon{file-alt}} \\

MathOdyssey \cite{fang2025mathodyssey} & 2024.06.26 & \texttt{MathOdyssey/MathOdyssey} & -- 
& \href{https://huggingface.co/datasets/MathOdyssey/MathOdyssey}{\scriptsize\faExternalLink*}
& \href{https://www.nature.com/articles/s41597-025-05283-3}{\scriptsize\faIcon{file-alt}} \\

Omni-MATH \cite{gao2024omnimath} & 2024.09.14 & \texttt{KbsdJames/Omni-MATH} & -- 
& \href{https://huggingface.co/datasets/KbsdJames/Omni-MATH}{\scriptsize\faExternalLink*}
& \href{https://arxiv.org/abs/2410.07985}{\scriptsize\faIcon{file-alt}} \\

HARP \cite{yue2024harp} & 2024.12.11 & \texttt{HARP (Released only on GitHub)} & -- 
& \href{https://huggingface.co/datasets/GITHUB/HARP}{\scriptsize\faExternalLink*}
& \href{https://arxiv.org/abs/2412.08819}{\scriptsize\faIcon{file-alt}} \\

HLE-Math \cite{HLE2026} & 2025.01.24 
& \texttt{cais/hle} & -- 
& \href{https://huggingface.co/datasets/cais/hle}{\scriptsize\faExternalLink*} 
& \href{http://dx.doi.org/10.1038/s41586-025-09962-4}{\scriptsize\faIcon{file-alt}} \\

BeyondAIME \cite{seed2025beyondaime} & 2025.06.17 & \texttt{ByteDance-Seed/BeyondAIME} & -- &
\href{https://huggingface.co/datasets/ByteDance-Seed/BeyondAIME}{\scriptsize\faExternalLink*} 
& \href{https://arxiv.org/abs/2504.13914v2}{\scriptsize\faIcon{file-alt}} \\

AMO-Bench \cite{an2025amobench} & 2025.10.29 & \texttt{meituan-longcat/AMO-Bench} & -- &
\href{https://huggingface.co/datasets/meituan-longcat/AMO-Bench}{\scriptsize\faExternalLink*} 
& \href{https://arxiv.org/abs/2510.26768}{\scriptsize\faIcon{file-alt}} \\

IMO-AnswerBench \cite{luong-etal-2025-imo-answerbench} & 2025.11.05 & \texttt{Hwilner/imo-answerbench} & -- 
& \href{https://huggingface.co/datasets/Hwilner/imo-answerbench}{\scriptsize\faExternalLink*} 
& \href{https://aclanthology.org/2025.emnlp-main.1794/}{\scriptsize\faIcon{file-alt}} \\

AMC23 \cite{amc23} & -- & \texttt{math-ai/amc23} & -- &
\href{https://huggingface.co/datasets/math-ai/amc23}{\scriptsize\faExternalLink*} 
& -- \\

AIME24 \cite{aime24} & -- & \texttt{math-ai/aime24} & -- &
\href{https://huggingface.co/datasets/math-ai/aime24}{\scriptsize\faExternalLink*} 
& -- \\

AIME25 \cite{aime25} & -- & \texttt{math-ai/aime25} & -- &
\href{https://huggingface.co/datasets/math-ai/aime25}{\scriptsize\faExternalLink*} 
& -- \\

\bottomrule
\end{tabular}
\caption{Data sources, RLVR datasets and test sets used in this work. For competition datasets (AMC23, AIME24, and AIME25), we cite the corresponding HuggingFace dataset pages since these datasets are primarily distributed through HuggingFace.}
\label{tab:dataset_details}
\end{table*}

%% file: table/source_info.tex
\begin{table*}[htbp]
\small
\centering
\setlength{\tabcolsep}{4pt}
\begin{tabular}{l r l l}
\toprule
\textbf{Source Label} & \textbf{Size} & \textbf{Problem Type} & \textbf{Merged Atomic Sources} \\
\midrule

\multicolumn{4}{l}{\textbf{Non-synthetic / Real-world Data}}\\[2pt]

\texttt{cn\_k12} & 342,217 & MWP &
\{cn\_k12\} \\

\texttt{olympiads} & 296,998 & Competition &
\{olympiads, olympiads\_ref\} \\

\texttt{aops\_forum} & 107,244 & Competition (Forum) &
\{aops\_forum\} \\

\texttt{stack\_exchange} & 97,025 & Forum &
\{deepmath\_stack\_exchange\} \\

\texttt{big\_math} & 32,227 & Mixed &
\{big\_math\_reformulated\} \\

\texttt{numina\_math1.5} & 26,821 & Mixed &
\{cn\_contest, inequalities, number\_theory\} \\


\texttt{still} & 18,175 & Mixed &
\{still1, still3\} \\

\texttt{gsm8k} & 14,869 & MWP &
\{gsm8k\} \\

\texttt{areal\_boba$^*$} & 14,838 & Mixed &
\{areal\_boba, areal\_boba\_synthetic\} \\


\texttt{amc\_aime} & 10,265 & Competition &
\{amc\_aime, aime19832023\} \\

\texttt{math} & 9,877 & Competition &
\{math, math(lila)\} \\

\texttt{dapo} & 5,294 & Mixed &
\{dapo\} \\

\texttt{num\_glue} & 4,925 & MWP &
\{num\_glue\} \\

\texttt{lila\_crawl} & 2,635 & MWP &
\{draw, asdiv\} \\

\texttt{basic\_arithmetic} & 1,829 & MWP &
\{addsub, simuleq, singleop, singleq, multiarith\} \\

\midrule
\multicolumn{4}{l}{\textbf{Synthetic / Weakly-Synthetic}}\\[2pt]

\texttt{synthetic\_math} & 149,520 & Synthetic &
\{synthetic\_math\} \\

\texttt{orca\_math} & 125,311 & Synthetic &
\{orca\_math\} \\

\texttt{lila\_synthetic} & 86,172 & Synthetic &
\{svamp, mathqa, deepmind\_mathematics, amps\} \\

\texttt{synthetic\_amc} & 83,407 & Synthetic &
\{synthetic\_amc\} \\

\texttt{meta\_math} & 11,011 & Synthetic &
\{meta\_math\} \\

\midrule
\multicolumn{4}{l}{\textbf{Special Handling / Filtered}}\\[2pt]

\texttt{test\_leak} & 6,093 & Competition &
\{omni\_math, math\_odyssey, harp, amc23\} \\

\texttt{no\_match} & 4,074 & Unknown &
\{N/A\} \\

\midrule
\textbf{Total} & 1,450,827 & -- & -- \\

\bottomrule
\end{tabular}
\caption{
Source labels and their merged atomic sources. 
\texttt{test\_leak} denotes datasets originally proposed as \textbf{evaluation benchmarks} without official train/dev/test splits, which may therefore be exposed to potential leakage risks depending on the training protocol.
Although a subset (e.g., \texttt{areal\_boba\_synthetic}) may contain weakly synthetic data based on our manual inspection, we conservatively categorize it as real-world data in this work.
}
\label{tab:data_source_detail}
\end{table*}

%% file: table/algo.tex
\begin{algorithm}[htbp]
\caption{Iterative Data Lineage Tracing}
\label{alg:lineage_tracing}
\begin{algorithmic}

\STATE \textbf{Input:} Initial dataset pool $\mathcal{P}$ sorted by time; thresholds $\tau,\delta$
\STATE \textbf{Output:} Lineage dictionary $\mathcal{L}[h]$
\STATE Initialize $\mathcal{L}\leftarrow\emptyset$, unmatched set $\mathcal{U}\leftarrow\emptyset$

\REPEAT

    \STATE \textbf{Stage 0: Data Collection and Canonicalization}
    \STATE collect candidate datasets $\mathcal{D}_{\text{new}}$
    \FOR{each dataset $D' \in \mathcal{D}_{\text{new}}$}
        \STATE canonicalize each instance into
        $d=(h,p,q,a,s,\text{id},t)$ via canonicalization and SHA1 hashing
        \STATE add processed dataset $D$ to $\mathcal{P}$
    \ENDFOR

    \STATE \textbf{Stage 1: Temporal Index Matching}
    \FOR{each instance $d=(h,p,q,a,s,\text{id},t)\in \mathcal{P}$ in chronological order}
        \IF{$h\in\mathcal{L}$}
            \STATE append $(\text{id},t,a,s)$ to $\mathcal{O}[h]$
        \ELSE
            \STATE create $\mathcal{L}[h]$ and initialize $\mathcal{O}[h]$ with $(\text{id},t,a,s)$
        \ENDIF
    \ENDFOR

    \STATE \textbf{Stage 2: Similarity-Based Matching}
    \STATE manually inspect unmatched records and apply rule set $R$ to construct $\mathcal{U}$
    \FOR{each instance $d\in\mathcal{U}$}
        \STATE find nearest prompt group $p\in\mathcal{L}[h]$ by cosine similarity
        \IF{$\cos(h,h')\geq\delta$}
            \STATE merge $\mathcal{O}[h]$ into $\mathcal{O}[h']$, remove key $h$ from $\mathcal{L}$, remove $h$ from $\mathcal{U}$
        \ENDIF
    \ENDFOR

    \STATE \textbf{Stage 3: Iterative Source Recovery}
    \IF{$|\mathcal{U}|>\tau$}
        \STATE manually identify candidate datasets, add $D_{\text{new}}$ to $\mathcal{P}$
        \STATE return Stage 1 and Stage 2
    \ENDIF

\UNTIL{$|\mathcal{U}|\leq\tau$}

\STATE \textbf{return} $\mathcal{L}[h]$

\end{algorithmic}
\end{algorithm}

%% file: table/case_study.tex
\begin{table*}[t]
\small
\centering
\setlength{\tabcolsep}{6pt}
\renewcommand{\arraystretch}{1.2}

\begin{tabular}{p{0.46\linewidth} p{0.46\linewidth}}
\toprule
\textbf{Training Set} & \textbf{Test Set} \\
\midrule

\multicolumn{2}{l}{\textbf{Case 1: \texttt{DeepScaleR-Preview-Dataset (omni\_math)} vs \texttt{math500} (sim = 0.9821)}} \\
\midrule

\begin{minipage}[t]{\linewidth}\cellsmall
\textbf{Prompt:}\\
Consider the function $z(x,$\showspace$y)$ describing the paraboloid\showspace
$z=(2$\showspace$ x-y)^{2}-2y^{2}-3$\showspace$ y$.
Archimedes and Brahmagupta are playing a game.
Archimedes first chooses $x$.
Afterwards, Brahmagupta chooses $y$.
Archimedes wishes to minimize $z$ while Brahmagupta wishes to maximize $z$.
Assuming that Brahmagupta will play optimally, what value of $x$ should Archimedes choose?

\vspace{2pt}
\textbf{Answer:} \texttt{\$-\textbackslash\textbackslash frac\{3\}\{8\}\$}

\end{minipage}
&
\begin{minipage}[t]{\linewidth}\cellsmall
\textbf{Prompt:}\\
Consider the function $z(x,y)$ describing the paraboloid\shownl

\showslash\showlb $z = (2x - y)^2 - 2y^2 - 3y.$\showslash\showrb
Archimedes and Brahmagupta are playing a game.\showspace\showspace
Archimedes first chooses $x.$\showspace\showspace
Afterwards, Brahmagupta chooses $y.$\showspace\showspace
Archimedes wishes to minimize $z$ while Brahmagupta wishes to maximize $z.$\showspace\showspace
Assuming that Brahmagupta will play optimally, what value of $x$ should Archimedes choose?

\vspace{2pt}
\textbf{Answer:} \texttt{\$-\textbackslash\textbackslash frac\{3\}\{8\}\$}

\end{minipage}

\\

\midrule

\multicolumn{2}{l}{\textbf{Case 2: \texttt{DAPO-Math-17k (areal\_boba)} vs \texttt{olympiad\_bench} (sim = 0.9818)}} \\
\midrule

\begin{minipage}[t]{\linewidth}\cellsmall
\textbf{Prompt:}\\
Let $\mathcal{A}$ denote the set of all polynomials in three variables $x, y, z$ with integer coefficients.
Let $\mathcal{B}$ denote the subset of $\mathcal{A}$ formed by all polynomials which can be expressed as:\shownl
\showlb
\shownl
$(x$\showspace$+$\showspace$y$\showspace$+$\showspace$z)P(x, y, z)$\showspace$+$\showspace
$(xy$\showspace$+$\showspace$yz$\showspace$+$\showspace$zx)$\\
$Q(x, y, z)$\showspace$+$\showspace
$xyzR(x, y, z)$ \shownl \showrb \shownl
where $P, Q, R \in \mathcal{A}$.
Find the smallest non-negative integer $n$ such that $x^i y^j z^k \in \mathcal{B}$ for all non-negative integers $i, j, k$ satisfying $i$\showspace$+$\showspace$j$\showspace$+$\showspace$k$\showspace \texttt{\showslash \red{geq}} \showspace$n$.

\vspace{2pt}
\textbf{Answer:} \texttt{4}

\end{minipage}
&
\begin{minipage}[t]{\linewidth}\cellsmall
\textbf{Prompt:}\\
Let $\mathcal{A}$ denote the set of all polynomials in three variables $x, y, z$ with integer coefficients.
Let $\mathcal{B}$ denote the subset of $\mathcal{A}$ formed by all polynomials which can be expressed as\shownl\shownl \texttt{\red{\$\$}}\shownl$(x+y+z)$\showspace$P(x, y, z)+(x$\showspace$y+y$\showspace$z+z$\showspace$x)$\showspace$Q(x, y, z)+x$\showspace$y$\showspace$z$\showspace$R(x, y, z)$\shownl\texttt{\red{\$\$}}\shownl\shownl with $P, Q, R \in \mathcal{A}$.
Find the smallest nonnegative integer $n$ such that $x^{\red{\{}i\red{\}}} y^{\red{\{}j\red{\}}} z^{\red{\{}k\red{\}}} \in \mathcal{B}$ for all nonnegative integers $i, j, k$ satisfying $i+j+k$ \showslash \texttt{\red{geqslant}} $n$.

\vspace{2pt}
\textbf{Answer:} \texttt{4}

\end{minipage}
\\

\midrule

\multicolumn{2}{l}{\textbf{Case 3: \texttt{DeepMath-103K (stack\_exchange)} vs \texttt{beyond\_aime} (sim = 0.9233)}} \\
\midrule

\begin{minipage}[t]{\linewidth}\cellsmall
\textbf{Prompt:}\\
\red{Let the r}eal numbers \(\showspace x_{1},\showspace x_{2},\showspace \cdots,\showspace x_{1997}\showspace\) satisfy the following conditions:\shownl
1.\showspace \(\showspace -\frac{1}{\sqrt{3}} \showspace \leq \showspace x_{i} \showspace \leq \showspace \sqrt{3} \showspace\)
for \(\showspace i = 1,\showspace 2,\showspace \cdots,\showspace 1997 \showspace\);\shownl
2.\showspace \(\showspace x_{1} \showspace + \showspace x_{2} \showspace + \showspace \cdots \showspace + \showspace x_{1997}
\showspace = \showspace -318 \showspace \sqrt{3} \showspace\).\shownl\shownl
\red{Find} the maximum value of \(\showspace x_{1}^{12} \showspace + \showspace x_{2}^{12} \showspace + \showspace \cdots
\showspace + \showspace x_{1997}^{12} \showspace\).

\vspace{2pt}
\textbf{Answer:} \texttt{189548}

\end{minipage}
&
\begin{minipage}[t]{\linewidth}\cellsmall
\textbf{Prompt:}\\
\red{R}eal numbers \(x_{1},x_{2},\cdots,x_{1997}\) satisfy the following \red{two} conditions:
(1)\(-\frac{1}{\sqrt{3}}\leq x_{i}\leq\sqrt{3}\)\showspace(\(i = 1,2,\cdots,1997\);\showspace
(2)\(x_{1}+x_{2}+\cdots +x_{1997}=-318\sqrt{3}\).\shownl
\red{Let \(M\) be} the maximal \red{possible} value of \(x_{1}^{12}+x_{2}^{12}+\cdots +x_{1997}^{12}\).
\red{Find the largest integer no more than \(M\).}

\vspace{2pt}
\textbf{Answer:} \texttt{189548}

\end{minipage}
\\

\midrule

\multicolumn{2}{l}{\textbf{Case 4: \texttt{Big-Math-RL-Verified-Processed (omni\_math)} vs \texttt{imo\_answerbench} (sim = 0.9002)}} \\
\midrule

\begin{minipage}[t]{\linewidth}\cellsmall
\textbf{Prompt:}\\
Sir Alex plays the following game on a row of 9 cells. Initially, all cells are empty.
In each move, Sir Alex is allowed to perform exactly one of the following two operations:\shownl

\redtexttt{[list=1]} \shownl \redtexttt{[*]} Choose any number of the form $2^j$, where $j$ is a non-negative integer, and put it into an empty cell.\shownl
\redtexttt{[*]} Choose two (not necessarily adjacent) cells with the same number in them; denote that number by $2^j$.
Replace the number in one of the cells with $2^{j+1}$ and erase the number in the other cell.\shownl
\redtexttt{[/list]}\shownl
At the end of the game, one cell contains \red{$2^n$, where $n$ is a given positive integer},
while the other cells are empty.
Determine the maximum number of moves that Sir Alex could have made\red{, in terms of $n$}.

\vspace{2pt}
\textbf{Answer:} \red{\texttt{$2 \sum_{i=0}^{8} \binom{n}{i} - 1$}}

\end{minipage}
&
\begin{minipage}[t]{\linewidth}\cellsmall
\textbf{Prompt:}\\
Sir Alex plays the following game on a row of 9 cells. Initially, all cells are empty.
In each move, Sir Alex is allowed to perform exactly one of the following two operations:\shownl\shownl
\red{(1)} Choose any number of the form $2^{j}$, where $j$ is a non-negative integer, and put it into an empty cell.\shownl\shownl
\red{(2)} Choose two (not necessarily adjacent) cells with the same number in them; denote that number by $2^{j}$.
Replace the number in one of the cells with $2^{j+1}$ and erase the number in the other cell.\shownl\shownl During the game, Sir Alex encounters a mysterious genie that grants him a wish.
\red{However, the genie warns Sir Alex that he can only make a limited number of moves.}
At the end of the game, one cell contains \red{the number $2^{40}$}, while the other cells are empty.
Determine the maximum number of moves that Sir Alex could have made.

\vspace{2pt}
\textbf{Answer:} \red{\texttt{200293447}}

\end{minipage}

\\

\bottomrule
\end{tabular}

\caption{
Representative high-similarity cases between evaluation benchmarks and training data.
Formatting differences are explicitly visualized using colored whitespace (\showspace),
newline tokens (\shownl), and display-math markers (\showlb/\showrb).
In both cases, the semantic content and final answers are identical,
while differences are purely superficial formatting artifacts.
}
\label{tab:case_study}
\end{table*}

%% file: table/hyperparameters.tex
\begin{table}[H]
  \centering
    \begin{tabular}{lc}
    \toprule
    \textbf{Hyper-parameter} & \textbf{Value} \\
    \midrule
    Learning Rate & 1e-6 \\
    Totel Steps & 500 \\
    Batch Size & 128 \\
    Mini Batch Size & 64 \\
    KL Loss Coefficient & 0.0 \\
    Clip Ratio & 0.2 \\
    Temperature & 1.0 \\
    Total Number of Rollouts & 8 \\
    Maximum Prompt Length & 4096 \\
    Maximum Response Length & 16384 \\
    \bottomrule
    \end{tabular}
  \caption{Full hyper-parameters for training.}
  \label{tab:hparam}
\end{table}

%% file: table/gpqa_results.tex
\begin{table}[H]
\centering
\small
\setlength{\tabcolsep}{2.5pt}
\renewcommand{\arraystretch}{1.08}
\begin{tabular}{lccc|c}
\toprule
  \textbf{Model} &  \textbf{Biology} &  \textbf{Physics} &  \textbf{Chemistry} &  \textbf{Overall}\\
\midrule

\textbf{\textit{Qwen3-1.7B-Base}}
& 32.9	& 28.5	& 30.1	& 29.7 \\

$\hookrightarrow$ DeepScaleR
& 43.4 & 29.7 & 34.1 & 33.1 \\

$\hookrightarrow$ DeepMath
& 48.7 & 39.2 & 42.5 & \textbf{41.7} \\

$\hookrightarrow$ OpenR1
& 34.2 & 27.3 & 33.9 & 31.1 \\

$\hookrightarrow$ DAPO
& 44.7 & 33.1 & 35.8 & 35.5 \\

$\hookrightarrow$ Skywork
& 36.8 & 35.2 & 38.2 & \underline{36.7} \\

\rowcolor{tblue} $\hookrightarrow$ DAPO++ (ours)
& 42.1 & 34.0 & 36.3 & 35.9 \\

\midrule
\addlinespace[2pt]

\textbf{\textit{Qwen3-8B-Base}}
& 60.5	& 45.9	& 37.9	& 43.6 \\

$\hookrightarrow$ DeepScaleR
& 56.6 & 47.1 & 33.9 & 41.8 \\

$\hookrightarrow$ DeepMath
& 60.5 & 58.7 & 33.3 & 47.0 \\

$\hookrightarrow$ OpenR1
& 61.8 & 39.8 & 40.9 & 42.4 \\

$\hookrightarrow$ DAPO
& 51.3 & 68.9 & 40.9 & \underline{54.0} \\

$\hookrightarrow$ Skywork
& 65.8 & 52.9 & 41.9 & 49.0 \\

\rowcolor{tblue} $\hookrightarrow$ DAPO++ (ours)
& 65.8 & 68.0 & 41.7 & \textbf{55.4} \\

\bottomrule
\end{tabular}
\caption{Mean@N results on the GPQA benchmark.}
\label{tab:gpqa_result}
\end{table}

%% file: table/source_meann_color.tex
\begin{table*}[htbp]
\centering
\small
\setlength{\tabcolsep}{4.5pt}
\renewcommand{\arraystretch}{1.08}
\begin{tabular}{lcccccccc|c}
\toprule
& \multicolumn{4}{c}{\textbf{Mean@4}} & \multicolumn{4}{c}{\textbf{Mean@32}} & \\
\cmidrule(lr){2-5}\cmidrule(lr){6-9}
\textbf{Source} & \textbf{Math500} & \textbf{Minerva} & \textbf{Olympiad} & \textbf{HLE} & \textbf{AMC23} & \textbf{AIME24} & \textbf{AIME25} & \textbf{AMO} & \textbf{Average} \\
\midrule

\textbf{\textit{Qwen3-1.7B-Base}} 
& 48.4 & 15.3 & 17.6 & 5.9
& 28.2 & 4.7 & 2.3 & 1.3
& 15.5 \\

\midrule

gsm8k
& \negheat{0}{25.9} & \negheat{0}{5.2} & \negheat{0}{9.1} & \negheat{0}{2.5}
& \negheat{0}{13.4} & \negheat{0}{2.5} & \negheat{0}{1.0} & \negheat{20}{0.9}
& \negheat{0}{7.6} \\

olympiads
& \posheat{22}{57.8} & \posheat{28}{22.3} & \posheat{38}{23.1} & \posheat{90}{6.1}
& \posheat{53}{35.1} & \posheat{87}{5.3} & \posheat{15}{5.2} & \posheat{58}{2.1}
& \posheat{36}{19.6} \\

areal\_boba
& \posheat{22}{57.8} & \posheat{19}{23.2} & \posheat{49}{22.1} & \posheat{57}{6.8}
& \posheat{40}{37.0} & \posheat{43}{7.3} & \negheat{77}{2.0} & \posheat{84}{1.6}
& \posheat{34}{19.7} \\

still3
& \second{\posheat{1}{60.4}} & \best{\posheat{0}{25.0}} & \posheat{31}{23.7} & \negheat{82}{5.3}
& \posheat{77}{31.6} & \posheat{61}{6.5} & \posheat{29}{4.7} & \negheat{0}{0.8}
& \posheat{34}{19.7} \\

synthetic\_amc
& \posheat{30}{56.9} & \posheat{22}{22.9} & \posheat{53}{21.8} & \second{\posheat{19}{7.6}}
& \posheat{40}{36.9} & \posheat{83}{5.5} & \posheat{65}{3.5} & \posheat{58}{2.1}
& \posheat{34}{19.7} \\

aops\_forum
& \posheat{31}{56.8} & \posheat{37}{21.4} & \posheat{51}{22.0} & \negheat{88}{5.5}
& \posheat{29}{38.5} & \posheat{48}{7.1} & \posheat{15}{5.2} & \posheat{74}{1.8}
& \posheat{33}{19.8} \\

stack\_exchange
& \posheat{18}{58.3} & \posheat{24}{22.7} & \posheat{53}{21.8} & \best{\posheat{0}{8.0}}
& \posheat{57}{34.5} & \posheat{63}{6.4} & \posheat{35}{4.5} & \posheat{42}{2.4}
& \posheat{33}{19.8} \\

basic\_arithmetic
& \posheat{17}{58.5} & \posheat{23}{22.8} & \posheat{35}{23.4} & \posheat{90}{6.1}
& \posheat{42}{36.7} & \posheat{67}{6.2} & \posheat{32}{4.6} & \posheat{68}{1.9}
& \posheat{30}{20.0} \\

num\_glue
& \posheat{14}{58.8} & \posheat{30}{22.1} & \posheat{40}{22.9} & \posheat{38}{7.2}
& \posheat{41}{36.8} & \posheat{85}{5.4} & \posheat{32}{4.6} & \posheat{63}{2.0}
& \posheat{30}{20.0} \\

still1
& \posheat{12}{59.0} & \posheat{28}{22.3} & \posheat{38}{23.1} & \negheat{76}{5.1}
& \posheat{28}{38.7} & \posheat{76}{5.8} & \posheat{68}{3.4} & \posheat{42}{2.4}
& \posheat{30}{20.0} \\

dapo
& \posheat{17}{58.5} & \posheat{36}{21.5} & \posheat{39}{23.0} & \negheat{88}{5.5}
& \posheat{21}{39.8} & \posheat{22}{8.3} & \posheat{38}{4.4} & \posheat{84}{1.6}
& \posheat{25}{20.3} \\

lila\_crawl
& \posheat{12}{59.0} & \posheat{26}{22.5} & \posheat{33}{23.6} & \posheat{86}{6.2}
& \posheat{35}{37.7} & \posheat{30}{7.9} & \posheat{35}{4.5} & \posheat{74}{1.8}
& \posheat{23}{20.4} \\

amc\_aime
& \posheat{15}{58.7} & \posheat{21}{23.0} & \posheat{33}{23.6} & \posheat{48}{7.0}
& \posheat{41}{36.8} & \posheat{20}{8.4} & \posheat{41}{4.3} & \posheat{68}{1.9}
& \posheat{22}{20.5} \\

lila\_synthetic
& \second{\posheat{1}{60.4}} & \posheat{12}{23.8} & \posheat{20}{24.7} & \negheat{53}{4.3}
& \posheat{43}{36.5} & \posheat{52}{6.9} & \best{\posheat{0}{5.7}} & \posheat{74}{1.8}
& \posheat{22}{20.5} \\

math
& \posheat{2}{60.3} & \posheat{11}{23.9} & \second{\posheat{8}{25.8}} & \negheat{65}{4.7}
& \posheat{23}{39.5} & \posheat{85}{5.4} & \posheat{56}{3.8} & \posheat{95}{1.4}
& \posheat{20}{20.6} \\

meta\_math
& \posheat{17}{58.5} & \posheat{8}{24.2} & \posheat{38}{23.1} & \posheat{29}{7.4}
& \posheat{38}{37.3} & \posheat{50}{7.0} & \second{\posheat{3}{5.6}} & \posheat{74}{1.8}
& \posheat{20}{20.6} \\

big\_math
& \posheat{7}{59.6} & \posheat{11}{23.9} & \posheat{17}{25.0} & \negheat{82}{5.3}
& \posheat{18}{40.2} & \posheat{54}{6.8} & \posheat{88}{2.7} & \posheat{74}{1.8}
& \posheat{19}{20.7} \\

numina\_math1.5
& \posheat{19}{58.2} & \posheat{18}{23.3} & \posheat{29}{23.9} & \posheat{57}{6.8}
& \posheat{14}{40.7} & \posheat{39}{7.5} & \posheat{71}{3.3} & \best{\posheat{0}{3.2}}
& \posheat{16}{20.9} \\

test\_leak
& \posheat{6}{59.8} & \posheat{14}{23.6} & \posheat{29}{23.9} & \negheat{88}{5.5}
& \second{\posheat{1}{42.7}} & \posheat{33}{7.8} & \posheat{44}{4.2} & \posheat{89}{1.5}
& \posheat{12}{21.1} \\

cn\_k12
& \posheat{9}{59.4} & \second{\posheat{2}{24.8}} & \posheat{28}{24.0} & \negheat{53}{4.3}
& \posheat{8}{41.6} & \best{\posheat{0}{9.3}} & \posheat{47}{4.1} & \posheat{58}{2.1}
& \posheat{11}{21.2} \\

orca\_math
& \posheat{4}{60.0} & \posheat{26}{22.5} & \posheat{17}{25.0} & \eqheat{5.9}
& \best{\posheat{0}{42.8}} & \posheat{22}{8.3} & \posheat{32}{4.6} & \posheat{74}{1.8}
& \second{\posheat{8}{21.4}} \\

synthetic\_math
& \best{\posheat{0}{60.5}} & \second{\posheat{2}{24.8}} & \best{\posheat{0}{26.5}} & \posheat{90}{6.1}
& \posheat{8}{41.6} & \second{\posheat{4}{9.1}} & \posheat{65}{3.5} & \second{\posheat{16}{2.9}}
& \best{\posheat{0}{21.9}} \\
\bottomrule
\end{tabular}
\caption{Mean@N results of RL checkpoints trained with different atomic source datasets under the SCA setting. Relative to Qwen3-1.7B-Base, improvements (blue) and degradations (red) are color-coded by intensity, respectively.}
\label{tab:mean_theta_vs_base}
\end{table*}

%% file: table/source_passn_color.tex
\begin{table*}[htbp]
\centering
\small
\setlength{\tabcolsep}{4.5pt}
\renewcommand{\arraystretch}{1.08}
\begin{tabular}{lcccccccc|c}
\toprule
& \multicolumn{4}{c}{\textbf{Pass@4}} & \multicolumn{4}{c}{\textbf{Pass@32}} & \\
\cmidrule(lr){2-5}\cmidrule(lr){6-9}

\textbf{Source} & \textbf{Math500} & \textbf{Minerva} & \textbf{Olympiad} & \textbf{HLE} & \textbf{AMC23} & \textbf{AIME24} & \textbf{AIME25} & \textbf{AMO} & \textbf{Average} \\
\midrule
\textbf{\textit{Qwen3-1.7B-Base}} 
& 66.6 & 28.7 & 28.7 & 17.2 & 80.0 & 33.3 & 23.3 & 8.0 & 35.7 \\
\midrule

gsm8k 
& \negheat{0}{42.2} & \negheat{0}{10.7} & \negheat{0}{18.7} & \negheat{0}{9.4}
& \negheat{0}{60.0} & \negheat{75}{30.0} & \negheat{67}{20.0} & \posheat{50}{\second{10.0}}
& \negheat{0}{25.1} \\

still3 
& \posheat{4}{\second{71.8}} & \posheat{41}{33.5} & \posheat{56}{33.3} & \negheat{30}{11.7}
& \negheat{75}{75.0} & \negheat{50}{26.7} & \negheat{34}{16.7} & \eqheat{8.0}
& \negheat{90}{34.6} \\

areal\_boba 
& \posheat{67}{68.4} & \posheat{41}{33.5} & \posheat{48}{34.1} & \negheat{60}{14.1}
& \negheat{88}{77.5} & \negheat{0}{20.0} & \negheat{67}{20.0} & \posheat{0}{\best{12.0}}
& \negheat{92}{34.9} \\

amc\_aime 
& \posheat{44}{69.6} & \posheat{32}{34.2} & \posheat{64}{32.4} & \negheat{80}{15.6}
& \negheat{88}{77.5} & \negheat{75}{30.0} & \negheat{100}{23.3} & \negheat{0}{6.0}
& \posheat{92}{36.1} \\

math 
& \posheat{37}{70.0} & \posheat{41}{33.5} & \posheat{36}{35.3} & \negheat{40}{12.5}
& \eqheat{80.0} & \negheat{25}{23.3} & \negheat{100}{23.3} & \posheat{0}{\best{12.0}}
& \posheat{90}{36.2} \\

numina\_math1.5 
& \posheat{30}{70.4} & \posheat{46}{33.1} & \posheat{50}{33.9} & \negheat{69}{14.8}
& \posheat{67}{82.5} & \negheat{75}{30.0} & \negheat{0}{13.3} & \posheat{0}{\best{12.0}}
& \posheat{88}{36.3} \\

synthetic\_amc 
& \posheat{15}{71.2} & \posheat{27}{34.6} & \posheat{61}{32.7} & \posheat{0}{\best{18.0}}
& \negheat{88}{77.5} & \negheat{50}{26.7} & \negheat{100}{23.3} & \posheat{50}{\second{10.0}}
& \posheat{80}{36.7} \\

still1 
& \posheat{44}{69.6} & \posheat{41}{33.5} & \posheat{62}{32.6} & \negheat{30}{11.7}
& \eqheat{80.0} & \negheat{100}{\second{33.3}} & \negheat{100}{23.3} & \posheat{0}{\best{12.0}}
& \posheat{75}{37.0} \\

orca\_math 
& \posheat{15}{71.2} & \posheat{64}{31.6} & \posheat{26}{36.3} & \negheat{30}{11.7}
& \eqheat{80.0} & \negheat{50}{26.7} & \posheat{50}{\second{30.0}} & \posheat{50}{\second{10.0}}
& \posheat{71}{37.2} \\

num\_glue 
& \posheat{33}{70.2} & \posheat{41}{33.5} & \posheat{56}{33.3} & \posheat{98}{\second{17.2}}
& \negheat{75}{75.0} & \negheat{75}{30.0} & \posheat{75}{26.7} & \posheat{0}{\best{12.0}}
& \posheat{71}{37.2} \\

stack\_exchange 
& \posheat{48}{69.4} & \posheat{54}{32.4} & \posheat{62}{32.6} & \posheat{0}{\best{18.0}}
& \negheat{88}{77.5} & \posheat{0}{\best{36.7}} & \negheat{100}{23.3} & \posheat{50}{\second{10.0}}
& \posheat{65}{37.5} \\

lila\_synthetic 
& \posheat{4}{\second{71.8}} & \posheat{32}{34.2} & \posheat{39}{35.0} & \negheat{30}{11.7}
& \negheat{75}{75.0} & \negheat{100}{\second{33.3}} & \posheat{50}{\second{30.0}} & \posheat{50}{\second{10.0}}
& \posheat{63}{37.6} \\

aops\_forum 
& \posheat{15}{71.2} & \posheat{73}{30.9} & \posheat{48}{34.1} & \negheat{60}{14.1}
& \eqheat{80.0} & \posheat{0}{\best{36.7}} & \posheat{75}{26.7} & \eqheat{8.0}
& \posheat{61}{37.7} \\

dapo 
& \posheat{15}{71.2} & \posheat{51}{32.7} & \posheat{30}{35.9} & \negheat{69}{14.8}
& \eqheat{80.0} & \negheat{50}{26.7} & \posheat{50}{\second{30.0}} & \posheat{50}{\second{10.0}}
& \posheat{61}{37.7} \\

big\_math 
& \posheat{11}{71.4} & \posheat{23}{34.9} & \posheat{22}{\second{36.7}} & \negheat{60}{14.1}
& \eqheat{80.0} & \negheat{75}{30.0} & \negheat{100}{23.3} & \posheat{0}{\best{12.0}}
& \posheat{59}{37.8} \\

meta\_math 
& \posheat{56}{69.0} & \posheat{18}{\second{35.3}} & \posheat{35}{35.4} & \posheat{0}{\best{18.0}}
& \negheat{88}{77.5} & \negheat{100}{\second{33.3}} & \posheat{75}{26.7} & \eqheat{8.0}
& \posheat{57}{37.9} \\

test\_leak 
& \posheat{37}{70.0} & \posheat{37}{33.8} & \posheat{45}{34.4} & \negheat{50}{13.3}
& \posheat{33}{\second{85.0}} & \negheat{75}{30.0} & \posheat{75}{26.7} & \posheat{50}{\second{10.0}}
& \posheat{57}{37.9} \\

synthetic\_math 
& \posheat{19}{71.0} & \posheat{41}{33.5} & \posheat{0}{\best{39.0}} & \negheat{80}{15.6}
& \negheat{62}{72.5} & \negheat{100}{\second{33.3}} & \posheat{75}{26.7} & \posheat{0}{\best{12.0}}
& \posheat{57}{37.9} \\

basic\_arithmetic 
& \posheat{41}{69.8} & \posheat{37}{33.8} & \posheat{30}{35.9} & \negheat{69}{14.8}
& \posheat{33}{\second{85.0}} & \negheat{50}{26.7} & \posheat{50}{\second{30.0}} & \eqheat{8.0}
& \posheat{55}{38.0} \\

lila\_crawl 
& \posheat{26}{70.6} & \posheat{54}{32.4} & \posheat{33}{35.6} & \negheat{90}{16.4}
& \posheat{67}{82.5} & \negheat{100}{\second{33.3}} & \negheat{100}{23.3} & \posheat{50}{\second{10.0}}
& \posheat{55}{38.0} \\

olympiads 
& \posheat{26}{70.6} & \posheat{27}{34.6} & \posheat{35}{35.4} & \negheat{80}{15.6}
& \posheat{33}{\second{85.0}} & \negheat{75}{30.0} & \posheat{50}{\second{30.0}} & \posheat{50}{\second{10.0}}
& \posheat{37}{\second{38.9}} \\

cn\_k12 
& \posheat{0}{\best{72.0}} & \posheat{0}{\best{36.8}} & \posheat{23}{36.6} & \negheat{30}{11.7}
& \posheat{0}{\best{87.5}} & \negheat{100}{\second{33.3}} & \posheat{0}{\best{36.7}} & \posheat{0}{\best{12.0}}
& \posheat{0}{\best{40.8}} \\
\bottomrule
\end{tabular}
\caption{Pass@N results of RL checkpoints trained with different atomic source datasets under the SCA setting. Relative to Qwen3-1.7B-Base, improvements (blue) and degradations (red) are color-coded by intensity, respectively.}
\label{tab:pass_theta_vs_base}
\end{table*}

%% file: table/hyperparameters_benchmark.tex
\begin{table}[H]
\centering
\small
\begin{tabular}{lcc}
\toprule
\textbf{Parameter} & \textbf{1B} & \textbf{8B} \\
\midrule

\multicolumn{3}{l}{\textbf{$S_{1a}$ coefficients: Verifiability-related scoring terms}} \\

$\alpha_{\mathrm{con}}$ & 0.50 & 0.50 \\
$\alpha_{\mathrm{mcq}}$ & 0.30 & 0.30 \\
$\beta_{\mathrm{mcq}}$ & 0.50 & 0.50 \\
$\alpha_{\mathrm{reuse}}$ & 0.20 & 0.20 \\

\midrule

\multicolumn{3}{l}{\textbf{$S_1$ weights: Static dataset quality composition}} \\

$w_{1a}$ & 0.20 & 0.20 \\
$w_{1b}$ & 0.55 & 0.50 \\
$w_{1c}$ & 0.25 & 0.30 \\

\midrule

\multicolumn{3}{l}{\textbf{$\alpha$ coefficients: SCA-based learnability utilities}} \\

$\alpha_{01}$ & $-0.3$ & $+1.5$ \\
$\alpha_{11}$ & $-0.5$ & $-0.8$ \\
$\alpha_{10}$ & $+1.5$ & $+0.5$ \\
$\alpha_{00}$ & $-1.5$ & $-0.8$ \\

\midrule

\multicolumn{3}{l}{\textbf{Composite $Q$ weights: Final dataset quality aggregation}} \\

$w_1$ & 0.50 & 0.35 \\
$w_2$ & 0.30 & 0.35 \\
$w_3$ & 0.20 & 0.30 \\

\bottomrule
\end{tabular}

\caption{Hyperparameters for the RLVR Dataset Quality Score $Q$.
Scale-dependent values are given for two representative model sizes;
intermediate values are obtained by linear interpolation in
$\log_{10}(M/\mathrm{1B})$.}

\label{tab:hyperparams_benchmark}
\end{table}